\documentclass[10pt,journal,compsoc]{IEEEtran}
	\usepackage{afterpage}

	\usepackage{cite}
	
\usepackage[latin9]{inputenc}
	\ifCLASSINFOpdf
	\else
	\fi
	\usepackage{enumitem}
	\usepackage{soul}
	
	\usepackage{cite}
	\usepackage{amssymb,amsmath,caption}
\usepackage[hang,flushmargin]{footmisc}
\usepackage{lipsum}
\makeatletter
\newcommand{\algorithmfootnote}[2][\footnotesize]{%
  \let\old@algocf@finish\@algocf@finish
  \def\@algocf@finish{\old@algocf@finish
    \leavevmode\rlap{\begin{minipage}{\linewidth}
    #1#2
    \end{minipage}}%
  }%
}
	\usepackage{soul,color}
	\usepackage{hyperref}

\usepackage{subfig}
	
	\usepackage{ragged2e}
	
	\usepackage{amsmath,varwidth,array,ragged2e}
	\usepackage{multirow}

	\usepackage{graphicx}
	
	\usepackage{tabularx}
\usepackage{booktabs}
\usepackage{longtable}
\usepackage{multirow}
\usepackage{colortbl}
\usepackage{caption}
\usepackage{hhline}
\usepackage{tabularx,colortbl}
\usepackage{mathtools}
\usepackage{eqnarray,amsmath}
\usepackage{amsmath}
\usepackage{hyperref} 
\usepackage{cite}
\usepackage{amsmath,amssymb,amsfonts}
\usepackage{algorithmic}
\usepackage{graphicx}
\usepackage{enumitem}
 \usepackage{subfig}
\usepackage{graphicx}
\usepackage{comment}
\usepackage{makecell}
\usepackage{wasysym}
\usepackage{mathrsfs}
\usepackage{amsthm}
\usepackage{bm}
\usepackage{soul}
\usepackage[export]{adjustbox}

\usepackage[ruled,linesnumbered]{algorithm2e}
\definecolor{seagreen}{rgb}{0.18, 0.55, 0.24}
\SetAlFnt{\small\color{black}\tt}
\LinesNumbered

\usepackage{amssymb}
\usepackage{pifont}

\makeatletter
\newcommand\notsotiny{\@setfontsize\notsotiny\@vipt\@viipt}
\makeatother

\makeatletter
\newcommand*\bigcdot{\mathpalette\bigcdot@{.5}}
\newcommand*\bigcdot@[2]{\mathbin{\vcenter{\hbox{\scalebox{#2}{$\m@th#1\bullet$}}}}}
\makeatother

\usepackage{amssymb}
\usepackage{pifont}

	\usepackage{mathtools}  
	\usepackage{mathtools}
	\usepackage{amsmath}
\usepackage{amsthm}
\usepackage{amssymb}

	\usepackage{array}
	
	\usepackage{amsmath}
\usepackage{amsthm}
\usepackage{amssymb}
\usepackage[export]{adjustbox}
\usepackage{verbatim}
\usepackage{graphicx}

	\usepackage{threeparttable}

	\usepackage[table]{xcolor}
	\usepackage{tikz}
	\usetikzlibrary{shapes,arrows}
	
	\tikzstyle{line}=[draw] 
	
	\usepackage{amsfonts}
\usepackage{mathrsfs}

		\usepackage[export]{adjustbox}
	\usepackage{verbatim}
	\usepackage{graphicx}
	\usepackage{caption}

	\usepackage{tabularx}
\usepackage{multicol, blindtext}
\usepackage{breqn}
\usepackage[T1]{fontenc}
\usepackage{multirow}
\usepackage{subfig}
	
	\usepackage{colortbl}
\usepackage{tabulary}
\usepackage{etoolbox}
\usepackage{colortbl}
\usepackage{bigstrut}
\usepackage{multirow}

\begin{document}
	%
\title{2L-3W: 2-Level 3-Way Hardware-Software Co-Verification for the Mapping of Deep Learning Architecture (DLA) onto FPGA Boards}	%
	\author{Tolulope A. Odetola, Katie M. Groves, and Syed Rafay Hasan
	\IEEEcompsocitemizethanks{\IEEEcompsocthanksitem Tolulope A. Odetola, Katie M. Groves, and Syed Rafay Hasan are with the Department of Electrical \& Computer
	Engineering, Tennessee Tech University, Cookeville, TN 38505 USA.\protect\\
	}
	

	}

\IEEEtitleabstractindextext{
	\begin{abstract}
	
 FPGAs have become a popular choice for deploying deep learning architectures (DLA). There are many researchers that have explored the deployment and mapping of DLA on FPGA. However, there has been a growing need to do design-time hardware-software co-verification of these deployments. To the best of our knowledge this is the first work that proposes a 2-Level 3-Way (2L-3W) hardware-software co-verification methodology and provides a step-by-step guide for the successful mapping, deployment and verification of DLA on FPGA boards. The 2-Level verification is to make sure the implementation in each stage (software and hardware) are following the desired behavior. The 3-Way co-verification provides a cross-paradigm (software, design and hardware) layer-by-layer parameter check to assure the correct implementation and mapping of the DLA onto FPGA boards. The proposed 2L-3W co-verification methodology has been evaluated over several test cases. In each case, the prediction and layer-by-layer output of the DLA deployed on PYNQ FPGA board (hardware) alongside with the intermediate design results of the layer-by-layer output of the DLA implemented on Vivado HLS and the prediction and layer-by-layer output of the software level (Caffe deep learning framework) are compared to obtain a layer-by-layer similarity score. The comparison is achieved using a completely automated Python script. The comparison provides a layer-by-layer similarity score that informs us the degree of success of the DLA mapping to the FPGA or help identify in design time the layer to be debugged in the case of unsuccessful mapping. We demonstrated our technique on LeNet DLA and Caffe inspired Cifar-10 DLA and the co-verification results yielded layer-by-layer similarity scores of 99\% accuracy.

	\end{abstract}
	
	
	}

	\maketitle

	\IEEEdisplaynontitleabstractindextext


	%

	\IEEEpeerreviewmaketitle

\section{Introduction}
Convolutional neural network (CNN), a well-known Deep Learning Architecture (DLA) evolved from artificial neural network, has been extensively applied to various applications, such as video surveillance, mobile robot vision, image search engine in data centers and so on \cite{zhang2015optimizing}. In general, deep learning uses a multi-layer neural network model to extract high-level features which are a combination of low-level abstractions to classify mutually exclusive properties of an image data \cite{tolu2}. This helps in finding the distributed data features, in order to solve complex problems in machine learning \cite{wang2017dlau}.

Due to the specific computation pattern of CNN, cloud computing have been employed to perform classification of deep learning models but this raises concerns of privacy ~\cite{baza2019b,baza2018blockchain,parksmarnet,baza2019blockchain,parkccnc,pazos2019privacy,baza2019detecting,Lightride,Andrew,shafee2019mimic,baza2015efficient,blockchainKey,firmware2}, security \cite{tolu1} and latency. General-purpose processors are also not efficient for CNN implementation and can hardly meet the performance requirement \cite{bacis2017pipelined}. Thus, various accelerators based on FPGA (Field Programmable Gate Array), GPU (Graphics Processing Unit), and even ASIC (Application Specific Integrated Circuits) design have been proposed to improve performance of CNN designs \cite{hailesellasie2019mulnet}. Among these approaches, FPGA based accelerators have attracted more attention because they have advantages of good performance, high energy efficiency, fast prototyping, and capability of reconfiguration \cite{zhang2015optimizing}.

To take advantage of what FPGA has to offer, several approaches like \cite{guo2017angel}, \cite{park2016fpga} and \cite{rastegari2016xnor} have been proposed to enable efficient optimizations for the deployment and successful mapping of DLAs onto FPGA boards. These optimizations help to reduce latency, conserve area and memory on the hardware (FPGA) \cite{zhang2017machine}. Majority of these mapping and optimizations are only validated at the point of final prediction and the measure of accuracy. Hence, layer-by-layer design time verification mechanism for DLA mapping to hardware from software paradigm to hardware paradigm has not been addressed. Verification is very crucial in hardware design as it accounts for about 80\% of modern hardware design time \cite{wang2009electronic}. 

Though, many researchers understand the crucial nature of verification in the design and mapping of DLA, but the mapping of DLA to FPGA board has unique phases from the software to the design and eventual mapping onto FPGA boards (hardware). Several approaches have been adopted to verify the workings and correctness of the DLA. Xiang et. al \cite{xiang2018output} proposes a software simulation based approach to verify the correctness of multilayer neural networks by measuring the maximum sensitivity of the layers of the network. Similarly, Dwarakanath et. al \cite{dwarakanath2018identifying} proposes a software based approach to verify the correctness of image classifiers by building relationships between subsequent layer-by-layer outputs corresponding to different inputs. These verification approaches are limited to software level layer-by-layer output and the accuracy of the final prediction. They do not provide a means of verifying the implementation correctness of the mapping of DLAs on hardware.

The verification approach verifies to the layer-by-layer output and the accuracy of the final prediction only. This approach does not take into consideration an approach that can be applied to the mapping of DLAs to FPGA boards.

 Other approaches that focus on the mapping of DLA onto FPGA boards involve the process of hardware-software co-design. Guo et. al \cite{guo2017angel} proposes a design flow for mapping CNNs onto embedded FPGA using data quantization to reduce the bit-width of CNN models without compromising much on accuracy.  Similarly, Jiandong et. al \cite{mu2018collaborative} proposes a collaborative framework to optimize the OpenCL based CNN design. These co-design techniques can only validate the correctness of the implementation based on the accuracy of the final prediction.

Very recently, Cong et. al \cite{hao2019fpga} proposes a time saving co-design methodology that simultaneously searches possible design options to auto-generate efficient DNNs optimized for FPGA deployment. This design approach is all automatic from the software to the hardware deployment. However, it only validates the design based on the accuracy of prediction of the model.

One shortcoming in the existing literature in the process of mapping DLAs to FPGAs is their inability to show a complete hardware-software co-verification schemes of the hardware implementation against its counterpart software-based DLA. Some of the above mentioned approaches \cite{guo2017angel}, \cite{mu2018collaborative} and \cite{hao2019fpga} only have means of validating or debugging the deployment at the stage of final prediction while others \cite{xiang2018output} and \cite{dwarakanath2018identifying} show means of verifying the DLA in a software environment. During the mapping of DLA to FPGA, if the prediction in design is wrong or does not correspond to the software implementation, the traditional approaches of verification are not able to analyze layer-by-layer feature values of DLA in design time. In this paper, we work with the premise that for a sustainable DLA design environment co-verification at the three stages of design (software stage, hardware design stage and hardware deployment stage) is crucial. Hence, there is a need for a methodology for complete hardware-software co-verification of DLA that readily shows the step-by-step and end-to-end process of deployment and verification of the inference phase (the forward propagation path) of the DLAs over FPGA boards. To the best of the authors knowledge, no such methodology exists so far in the literature.

In this paper, we propose a 2-Level 3-Way (2L-3W) coherent hardware-software co-verification approach. Our 2-level verification approach is divided into software inference level and hardware inference level of the DLA. Our 3-way co-verification technique provides a means if assuring that the software design, hardware design and hardware mapping of the DLA are coherent and correctly implemented.

The following are the contributions of this work:
\begin{itemize}
\item A step-by-step and end-to-end methodology for the mapping of DLAs onto FPGA boards
\item A 2-level verification approach to ensure the implementation correctness of a designed DLA in both software and hardware
\item A 3-way layer-by-layer co-verification technique that ensures successful mapping of DLA to FPGA boards
\end{itemize}

The remainder of this paper is organized as follows: Section II provides some preliminary information. Section III discusses the proposed methodology of hardware-software co-verification of DLA. Section IV describes the experimental validation of the co-verification methodology setup on Xilinx PYNQ FPGA board. Section V shows the results and lessons learned. Section VI discusses the related work. Section VII compares the co-verification methodology with different approaches. Section VIII concludes the paper.

\section{Preliminaries}
\label{Background}
In order to understand this paper, we are providing information about some of the concepts used in this paper.
\subsection{High Level Synthesis (HLS)}
Hardware accelerators like FPGA provides a means to achieve moderate level performance with low power consumption, massive memory parallelism and short time to market\cite{park2017optimizing}. To ensure proper deployment of DLA on FPGA, hardware-software co-verification is essential. Hardware-software co-verification helps to ensure the behavior of the embedded system software is consistent with the hardware design.

\begin{figure*}[t]
\setlength{\abovecaptionskip}{0mm}   
\setlength{\belowcaptionskip}{0mm}   
\centering
\includegraphics[scale=0.45]{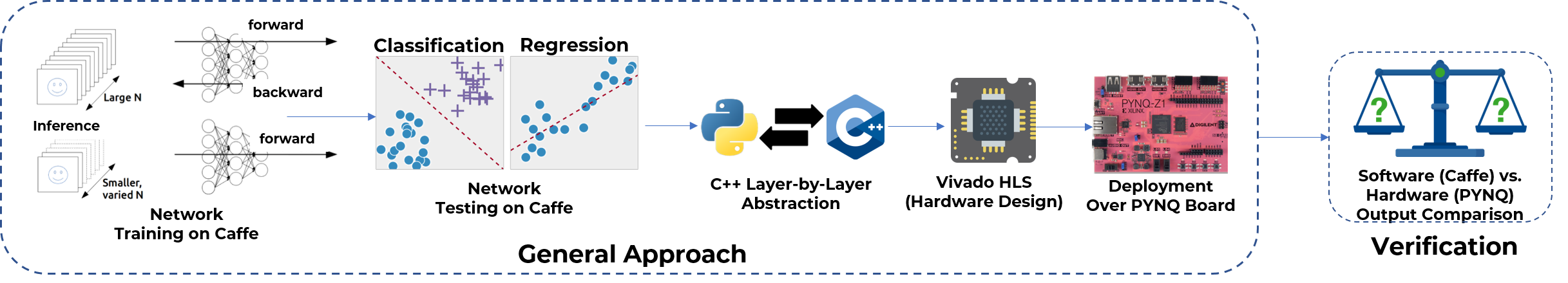}
\caption{General Deep Learning Approach.}
\vspace{0mm}
\label{fig:DL}
\end{figure*}

Hardware design using Hardware Descriptive Languages (HDL) can be time consuming and difficult to debug and verify \cite{o2014xilinx}. High Level Synthesis (HLS) offers flexibility by utilizing C/C++ code with a set of derivatives to automatically generate HDL for hardware implementation on FPGA. HLS provides a means of converting C/C++ code (High Level Languages) to HDL like VHDL or Verilog.

\subsection{Deep Learning Framework: Caffe}
In this paper, Caffe deep learning framework is adopted because of its popularity, support and easy-to-use interface. It is easy to experiment with popular pre-trained models \cite{lacey2016deep}. Caffe provides toolkits for training, fine-tuning and the deploying DLA \cite{jia2014caffe}. In Caffe, the DLA is designed and configured using prototxt files prior to training. After training, Caffe generates a caffemodel file containing the trained parameters (weights and biases) of the DLA. The parameters in the caffemodel file can be accessed through using Python libraries.

\subsection{Network Surgery}
DLA tend to have stacked layers. Each layer contains learnable parameters (weight and biases) \cite{guo2016dynamic}. For proper replication and deployment of the DLA on hardware, access and extraction of these learnable parameters are needed. During the inference phase, network surgery gives access to the output of each layer when an unseen data is passed through the DLA. The output of a layer is called Blob.  Network surgery allows access and extraction of the DLA parameters and Blobs.

\subsection{Chosen FPGA Board: PYNQ}
The hardware environment chosen is the PYNQ-Z1 FPGA \cite{bbb2017}. PYNQ-Z1 is built upon Xilinx ZYNQ SoC technology and is used to develop applications for ZYNQ-7000 based devices \cite{janssen2017dynamic}. The PYNQ platform offers designers the privilege of exploiting the programmable logic of the FPGA board from a Python environment\cite{janssen2017dynamic}. Xilinx provides Python packages that facilitates the interaction with hardware modules using overlays.

\subsection{Python Overlay}
Overlays, or hardware libraries, are configurable FPGA designs capable of extending user application from the ZYNQ processor of a PYNQ board into the programmable Logic. Overlays can be loaded to the FPGA dynamically like a software library. PYNQ overlays are created by hardware designers, and wrapped with PYNQ's Python Overlay API. This allows Python interface to program and control specialized hardware overlays \cite{pynq2019}.

\subsection{Data Transfer: AXI Direct Memory Access (AXI DMA)}
AXI DMA transfers data  between memory and AXI4-Stream-type target peripherals \cite{jeff2014}.  AXI DMA in Vivado provides high-bandwidth direct memory access between an AXI4 memory-mapped and an AXI4-Stream ports on IPs (Intellectual Property) interfaces \cite{Xil2019}. PYNQ supports the AXI central DMA IP with the PYNQ DMA class \cite{pynq2019}. DMA can be used for high performance burst transfers between Processing System (PS) DRAM and the Programmable Logic (PL). It helps to offload data from the Central Processing Unit (CPU) in processor-based systems \cite{Xil2019}.  AXI DMA data movement between system memory and stream target is through the AXI4 Read Master to AXI4 memory-mapped to stream (MM2S) Master, and AXI stream to memory-mapped (S2MM) Slave to AXI4 Write Master.

\begin{figure*}[p]
\setlength{\abovecaptionskip}{0mm}   
\setlength{\belowcaptionskip}{0mm}   
\centering
\includegraphics[scale=0.42]{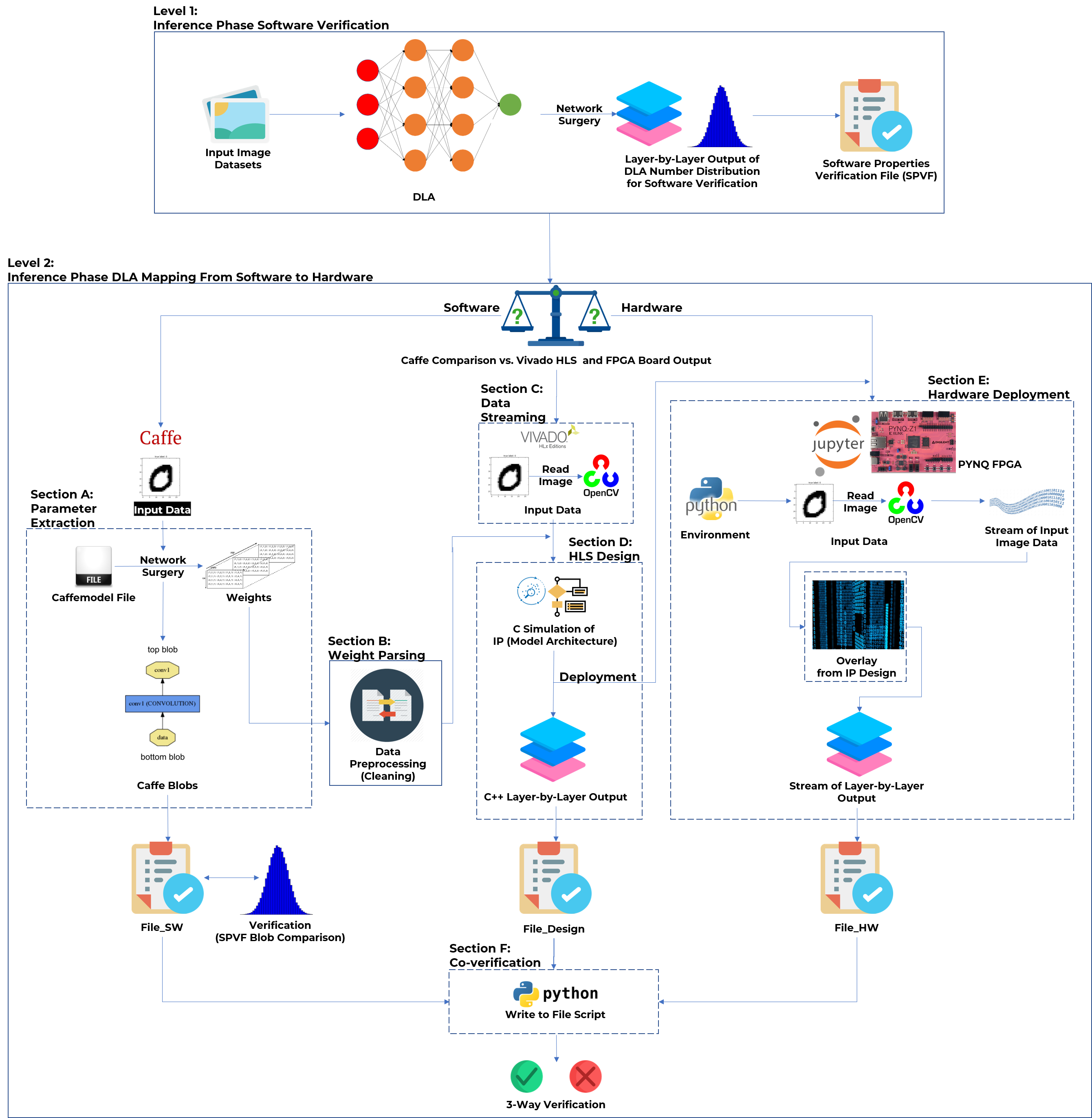}
\caption{Verification Approach.}
\vspace{0mm}
\label{fig:DL1}
\end{figure*}
\subsection{General Deep Learning Approach}
Fig. \ref{fig:DL} shows the general end-to-end approach from the training of a DLA to its deployment on FPGA board. This include the following steps:
\begin{itemize}
	\item \textit{Network Training:} This stage takes place after the DLA has been designed. The training process is where the best sets of parameters that maximizes a DLA's accuracy is determined by leveraging on gradient descent (back propagation). Training involves a number of forwared and backward propagation based on the number of iterations specified in the model design. Network training is done with CPUs or GPUs on different software frameworks like caffe, tensoflow and so on.
    \item \textit{Network Testing:} This stage is also referred to the inference stage. The trained model is used to classify unseen data and predict a result with a degree of accuracy.
    \item \textit{C++ Layer-by-Layer Abstraction:} Model design, training and testing are usually done in Python environment. For hardware design, the model design is converted from prototxt syntax adopted in Caffe (which is utilized using Python libraries in model training)to C++ syntax used in hardware design. In this stage, every layer is designed in C++ as stipulated in the model design in the prototxt. All conditions in terms of layer outputs, kernel sizes, stride sizes and so on for each respective layer is obeyed during this conversion.
    \item \textit{Vivado HLS (Hardware Design):} Vivado HLS provides an environment for the simulation and synthesis of the C++ code of the model design. After successful synthesis, Vivado HLS allows for IP generation of the model design.
    \item \textit{Deployment over FPGA (PYNQ Board):} In this stage, the IP generated from Vivado HLS is converted to bitstreams and deployed on the FPGA board.
\end{itemize}
The verification part shown on the right hand side of Fig. \ref{fig:DL} is something outside the realm of the general deep learning deployment methodology.

\section{Hardware-Software Co-Verification of DLA Inference Phase}
\label{methodology}
In this paper we are proposing a novel 2L-3W hardware-software co-verification concept for DLA deployment on FPGA boards. In order to achieve this, Caffe software framework is utilized for the software implementation (training and testing) and Vivado HLS for the hardware design synthesis. Finally, our approach uses Xilinx PYNQ FPGA board for hardware implementation

Prior to explaining our proposed co-verification appoach it is worthy to note that we collect the trained model apriori using Caffe deep learning framework. This trained model is called Caffemodel file in Caffe framework. Furthermore, we design the feed forward path of the trained DLA using Vivado HLS.

Fig. \ref{fig:DL1} shows the different levels and sections of the co-verification methodology. These sections are discussed as follows:

\subsection{Level 1: Inference Phase Software Verification}
This is the first level of the proposed 2L-3W co-verification methodology. In this phase, the trained DLA is collected. As shown in Fig. \ref{fig:DL1}, the image dataset (correctly predicted by the trained DLA) used in training the DLA is passed through the trained model. Network surgery is used to get the Blobs (layer-by-layer output features) of each layer of the DLA. These Blobs are obtained and used to investigate the numerical distribution of each respective layer Blob. Statistical properties like the range, minimum, maximum and standard deviation of each Blob is also collected over a given number of training image set and generalized and written to the Software Properties Verification File (SPVF) as shown in Fig. \ref{fig:DL1}. The SPVF file contains boundary values of each element in the Blob of each layer. This forms a benchmark for comparing the numerical distribution and statistical properties of Blobs of subsequent images (test images) that is passed through the model. This serves as the Inference Phase Software Verification as shown in Fig. \ref{fig:DL1}.

\subsection{Level 2: Inference Phase DLA Mapping From Software to Hardware}
This is the second level of the proposed 2L-3W co-verification methodology. This level is divided into 6 sections. After the software verification is done, the DLA is mapped to FPGA and test images are used to verify the implementation correctness of mapping the DLA from software to hardware (FPGA).

\subsubsection{Section A: Parameter Extraction Using Network Surgery}
This stage of the co-verification methodology is shown as Section A in Fig. \ref{fig:DL1}. This section shows that the trained model (i.e. Caffemodel file) parameters is obtained using a Caffe function called Network Surgery. Here, unseen data (data not used in the training phase) shown as input data in Fig. \ref{fig:DL1} is passed through the parameters of the trained Caffemodel file. This stage is carried out in the software environment. The prediction and the layer-by-layer output (Blobs) is extracted and written to a specified file called File\_SW as shown in Fig. \ref{fig:DL1}. The numerical distribution and statistical properties of the Blobs written to File\_SW  is then compared and validated with the properties written to the SPVF generated in level 1. Line 1 to 7 of Algorithm \ref{alg:MYALG} shows what actions needs to be taken if the DLA requirement is not met.
\subsubsection{Section B: Parsing of Weights from Caffe to Vivado HLS}
The parameters of the Caffemodel are obtained in Python and passed through a data cleaning process to convert the layer-by-layer parameters to be compatible with C++ syntax required for Vivado HLS. This converted parameters are then incorporated for Vivado HLS synthesis of hardware design. This is further explained in the simulation of HLS design section (Setion D). Line 8 to 11 of Algorithm \ref{alg:MYALG} summarizes this section

\subsubsection{Section C: Streaming of Input Data to the Hardware Design}
This section is in the hardware design stage. Here, the input data (unseen data), that is used in Section A is read using OpenCv C++ library and converted to a stream of data using HLS stream library. The stream of data is passed down to the designed IP in the HLS design (Section D). Line 12 of Algorithm \ref{alg:MYALG} summarizes this section.
\subsubsection{Section D: Simulation of HLS Design to Provide Layer-by-layer Output}
This section is shown in Fig. \ref{fig:DL1} as section D. This section assumes that the C++ adaptation of each layer of the DLA has been completed to form the IP in Vivado HLS. The weights obtained using the parsing of weights of the Caffemodel file (section B) is imported and merged appropriately with the IP. Unseen image data is read using OpenCV (as shown in section C) and is converted to a stream of data. This stream of data is passed as an input to the designed IP. After simulation of the IP, the layer-by-layer output and the prediction is written to a specified file (denoted as File\_Design) as shown in Fig. \ref{fig:DL1}. Line 13 to 16 of Algorithm \ref{alg:MYALG} shows what actions needs to be taken if the verification does not meet the requirement.

\subsubsection{Section E: Hardware Deployment and On-board Verification}
\begin{figure*}[p]
\setlength{\abovecaptionskip}{0mm}   
\setlength{\belowcaptionskip}{0mm}   
\centering
\includegraphics[scale=0.6]{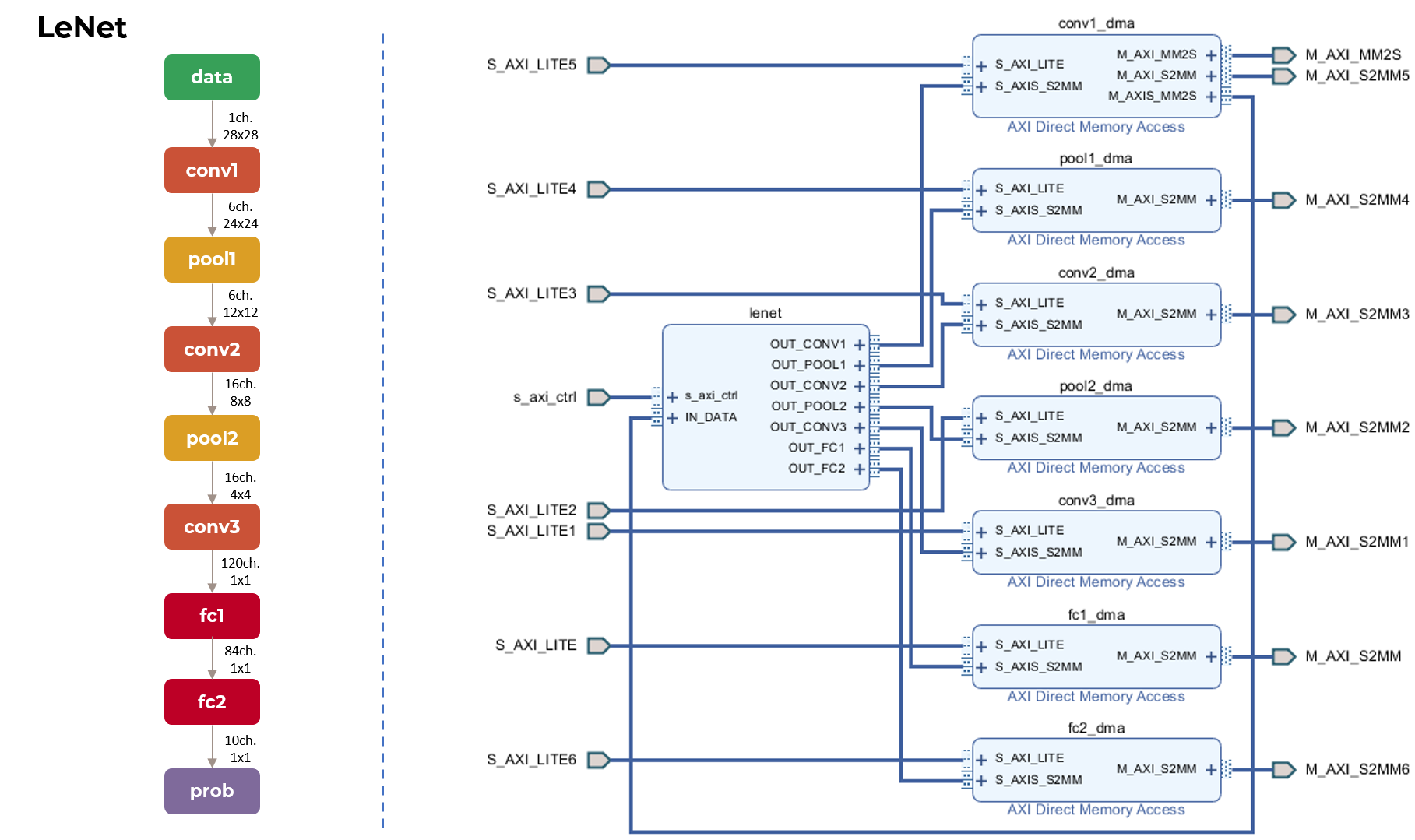}
\caption{LeNet DLA and hardware configuration for the output of each layer on FPGA}
\vspace{0mm}
\label{dma}

\includegraphics[scale=0.55]{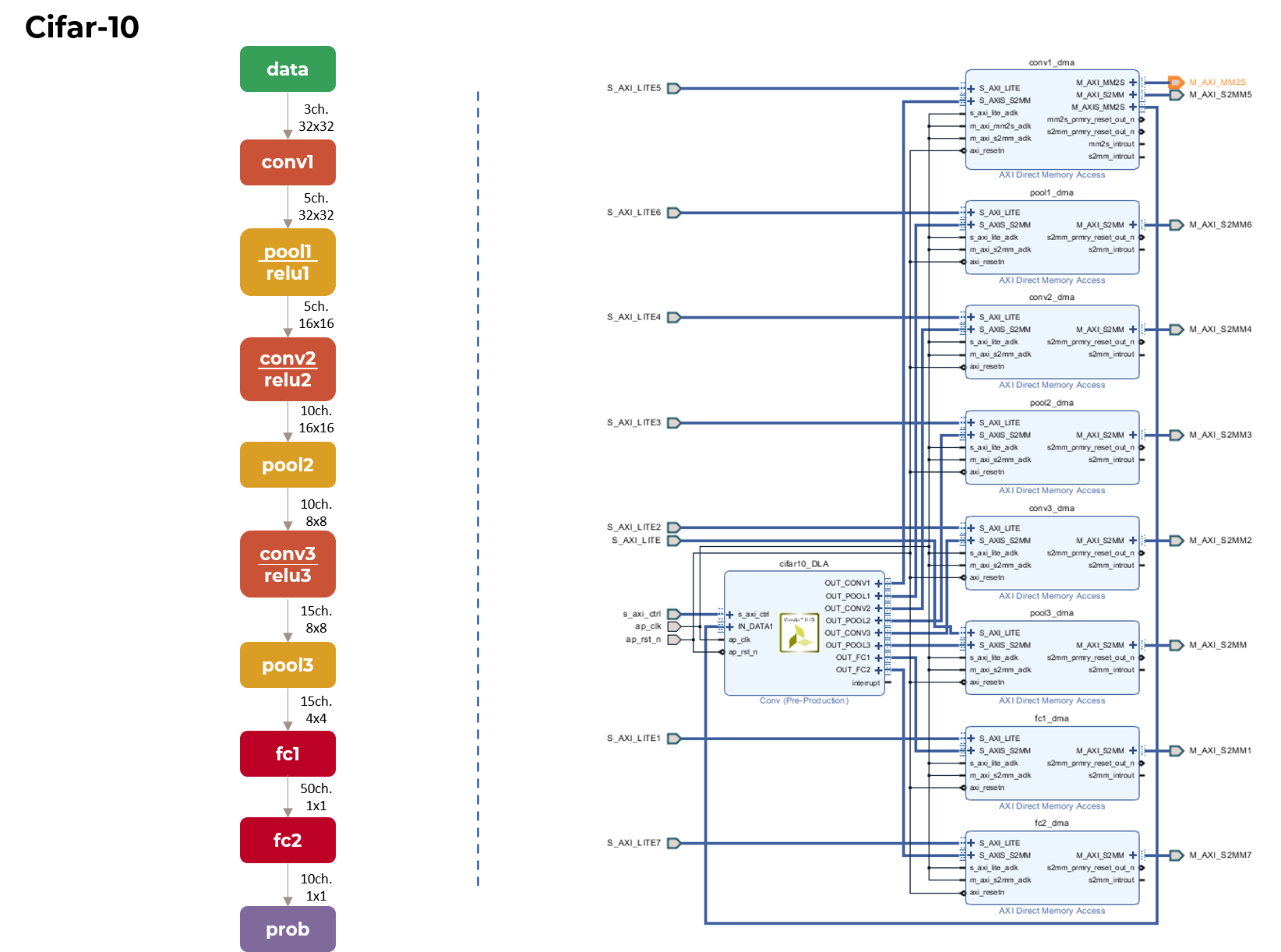}
\caption{Cifar-10 DLA and hardware configuration for the output of each layer on FPGA}
\vspace{0mm}
\label{dma2}
\end{figure*}

The generated IP is synthesized to obtain a bitstream and .tcl files in Xilinx Vivado environment as in the case of conventional design flow. These files are imported to the PYNQ board as an Overlay to be called in the Python environment. Special provisions are made to ensure the output of each stage of the DLA is compared against the output of simulation of HLS design (explained in Section D) and the software deep learning framework output (Caffe layer-by-layer output explained in Section A). Fig. \ref{dma} illustrates the comparison. It shows that all layers of the DLA are synthesized as a separate module. For example, the output of conv1\_dma (shown in the right hand side of Fig. \ref{dma}) corresponds to the output of conv1 layer in the software (shown in the left hand side of Fig. \ref{dma}). In order to store the values and automate the process we stored the output of each stage in Python environment in a separate file (denoted as File\_HW) as shown in Fig. \ref{fig:DL1}. Line 18 to 24 of Algorithm \ref{alg:MYALG} summarizes this section.

\subsubsection{Section F: Co-verification}
To automate our methodology, the output of all the three stages need to be compared seamlessly. In order to achieve this, a Python script is written that verifies the software verified layer-by-layer output of each stage that are stored in the File\_SW, File\_Design and File\_HW for our three-way verification approach. Line 25 to 28 of Algorithm \ref{alg:MYALG} summarizes this section and suggests possible actions if the verification does not meet the requirement.

\begin{algorithm}[!h]
\caption{2L-3W Hardware-software Co-verification Methodology}
\begin{algorithmic} [1]
\REQUIRE Design, Configure and training of Model
\STATE Testing of model on unseen data (D) in Caffe
\IF {Testing = Fails}
\STATE $Action: $ Retrain and re-design or re-configure model
\ELSE
\STATE Perform network surgery on model
\STATE Obtain layer-by-layer Blob and obtain the numerical distribution and generalized statistical properties (Range, Maximum, Minimum, Mean and Standard deviation) for correctly predicted images in training sets and write to SPVF
\ENDIF
\STATE Extract Blobs (layer-by-layer output) from testing (of unseen data), write to file (File\_SW).
\STATE Compare File\_SW with SPVF generated in 1evel 1
\STATE Extract parameters (weights and biases) of Model
\STATE Convert the parameters from Python syntax to C++
\STATE Implement C++ representation of each layer of the model design in Vivado HLS
\STATE Incorporate model design parameters with model design in Vivado HLS
\STATE Simulate model design with unseen data (D) used in model testing
\STATE Write layer-by-layer output of the result of simulation of the model in Vivado HLS to file (File\_Design)
\STATE Compare value-to-value of respective layer-by-layer output between Vivado HLS and Caffe
\IF {Vivado HLS Output != Caffe Output}
\STATE $Action: $ Redesign C++ algorithm and check for error using layer-by-layer output values
\ELSE
\STATE Generate IP from model design in Vivado HLS
\ENDIF
\STATE Configure IP in Vivado block design
\STATE Generate bit-stream
\STATE Deploy bit-stream on board
\STATE Import bit-stream in Python Overlay
\STATE Run bit-stream with unseen data (D) and write layer-by-layer output to file (File\_HW)
\STATE Perform hardware-software verification with results
\IF {FPGA Output != Vivado HLS Output or Caffe Output}
\STATE $Action: $ Redesign C++ algorithm and re-generate bitstream
\ELSE
\STATE End Deployment
\ENDIF
\end{algorithmic}
\label{alg:MYALG}
\end{algorithm}

\section{Experimental Validation of Hardware-Software Co-Verification}
\label{implementation}
To validate our methodology, we implemented 2 DLAs. The first DLA is LeNet and the other is Caffe-inspired Cifar-10 as shown in Figs. \ref{dma} and \ref{dma2}, respectively. Both DLAs are implemented on PYNQ Xilinx FPGA Board and the processes of implementation are for the most part the same. For the sake of vivid elaboration, our discussion in this section is explaining the process for LeNet DLA.

\begin{figure}[h]
\setlength{\abovecaptionskip}{0mm}   
\setlength{\belowcaptionskip}{0mm}   
\centering
\includegraphics[scale=0.4]{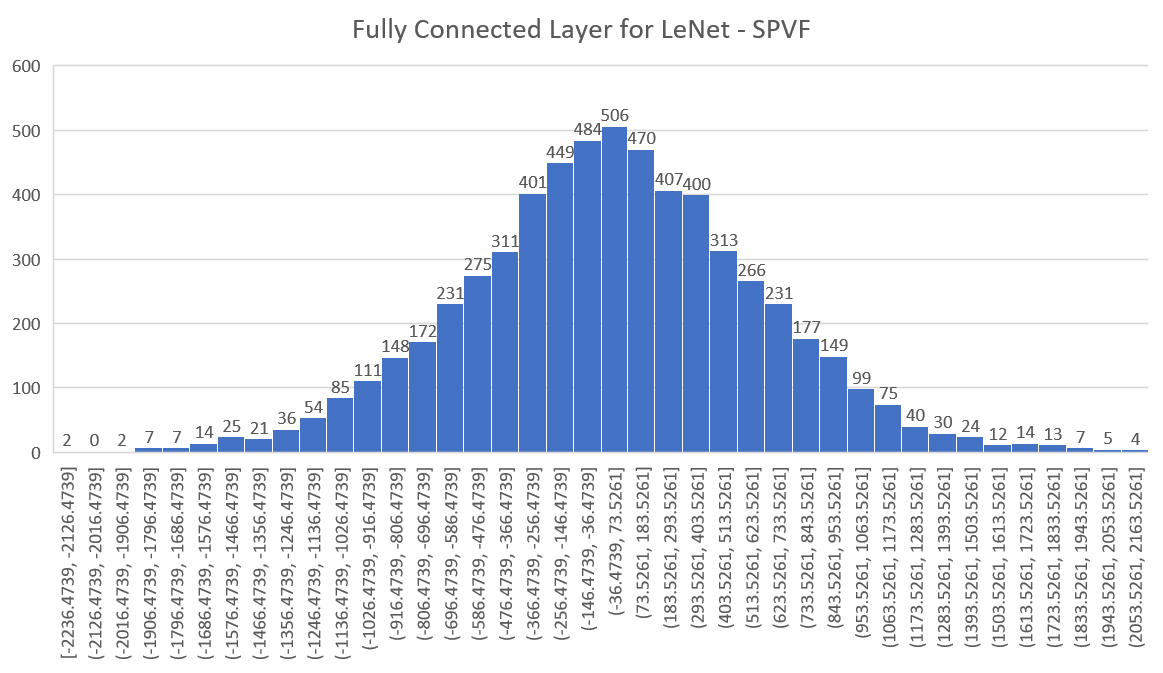}
\caption{Software Verification for Fully Connected Layer 1 for LeNet DLA}
\vspace{0mm}
\label{spvf}
\end{figure}

The LeNet DLA is designed and trained in Caffe. After training, several steps are taken to test the trained model and to validate the implementation correctness of the model. A total of 100 test images are passed through the model and it yields an accuracy of 97\%. After verifying the accuracy, 1000 correctly predicted images are passed through the LeNet DLA to obtain the numerical distribution of each respective Blob (Blob is explained in Section II) using network surgery. The mean, range, maximum, minimum and standard deviations are obtained and averaged over 1000 images to get generalized statistical properties of the each respective Blobs as shown in Fig. \ref{fig:DL1}. The boundary values (minimum and maximum of each element across all the chosen training imageset)for each element in the Blob is also obtained. These properties and boundary values are written to a specified file called SPVF. To illustrate this, Fig. \ref{spvf} shows the numerical distribution of outputs from the first fully connected layer of LeNet DLA. The Blobs of the first fully connected layer for unseen data is compared with this to verify it. The SPVF for the first fully connected layer shows the Blobs follow a Gaussian distribution. The same procedure is carried out for all the layers in the LeNet DLA. This concludes Level 1 of the 2L-3W hardware-software co-verification which is shown as ``Inference Phase Software Verification" in Fig. \ref{fig:DL1}.

\begin{figure*}[t]
\setlength{\abovecaptionskip}{0mm}   
\setlength{\belowcaptionskip}{0mm}   
\centering
\includegraphics[scale=0.36]{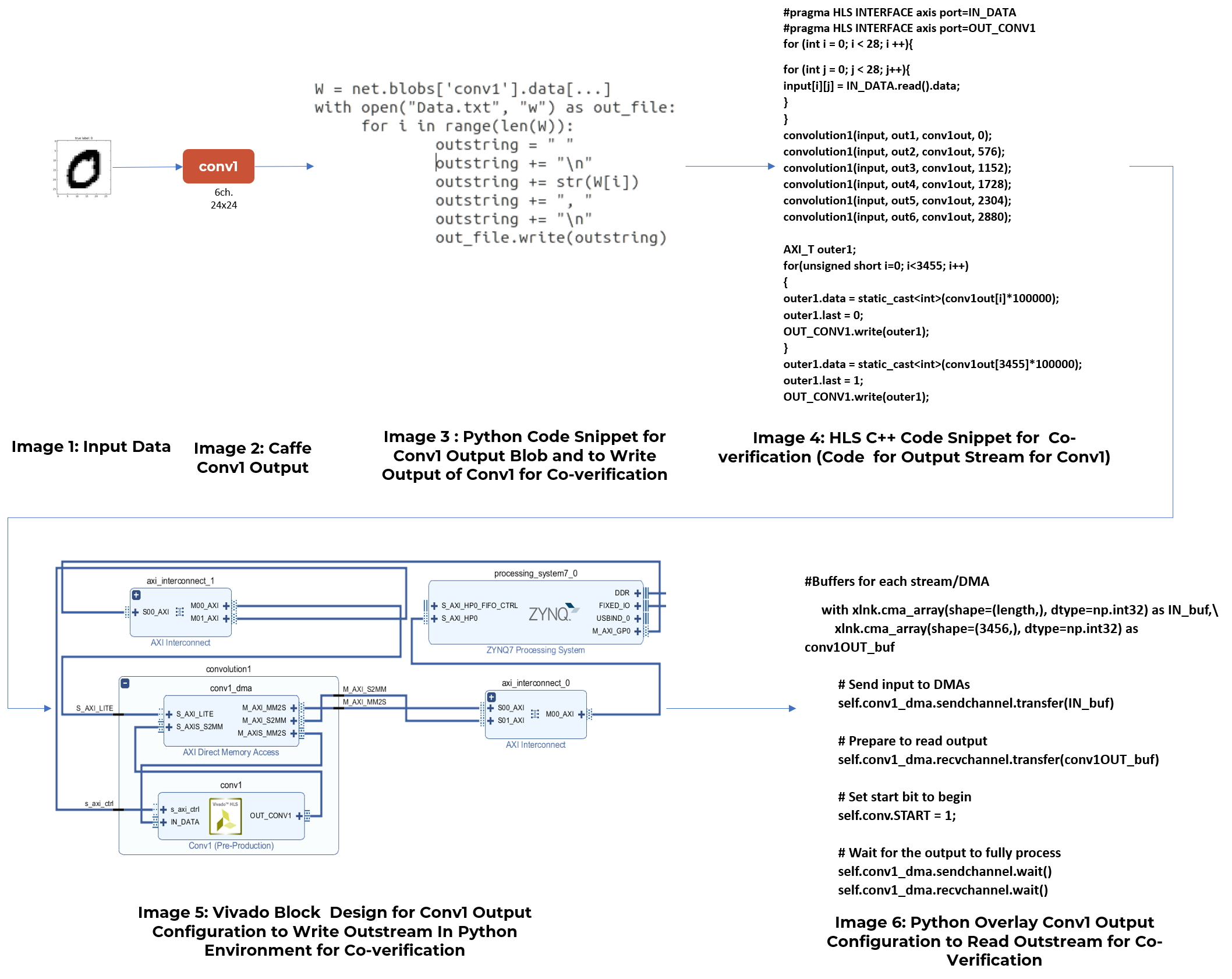}
\caption{Illustration of Conv1 for Caffe to HLS Design to Block Design to Python}
\vspace{0mm}
\label{sample}
\end{figure*}

For the second level of 2L-3W hardware-software co-verification, labelled as ``Level 2: Inference Phase DLA Mapping in Fig. \ref{fig:DL1}", an unseen image is passed through the trained DLA  and the Blobs of each layer are obtained using network surgery (explained in Section II) and written to a specified file denoted by File\_SW in Fig. \ref{fig:DL1}. The Blobs written to File\_SW is then verified with the SPVF. The element in the Blob of each layer is compared with the boundary values in the SPVF to verify them. The code snippet that allows the access to layer-by-layer output using network surgery for one of the DLA layers is shown in Image 3 of Fig. \ref{sample}. Following the proposed 2L-3W co-verification methodology in Fig. \ref{fig:DL1}, using Section B, the parameters (weights and biases) of the DLA are obtained and parsed into the HLS design. Each of the layers defined in the Caffe framework is also defined in Vivado HLS to maintain the same accuracy of prediction from Caffe to the PYNQ hardware. As shown in Section C in Fig. \ref{fig:DL1}, a stream of input data (same used in testing the Caffe model) is used in simulating the HLS design layers and the parsed parameters. The layer-by-layer output result of the simulation is written to a specified file denoted by File\_Design in Fig. \ref{fig:DL1}. The layer-by-layer output of the DLA written to File\_Design is then verified with the SPVF file. After successful verification, the DLA is optimized to fit the PYNQ FPGA board is then synthesized and packaged as an IP. Vivado HLS contains built in directives known as pragmas (shown in the first two lines of image 4 Fig. \ref{sample}) that specifies how the data is written to the IP (shown in Section C in Fig. \ref{fig:DL1}) and also how the data is read from the IP. The pragma used to allow data flow is called ``interface axis port". This axis port is important because this allows for an actual physical port of an AXI4-Stream to be used later in the block design. The AXI4-Streams ports allow for this implementation to Blobs from each layer to be viewed in the Python environment.

\begin{figure}[!h]
  \centering
\includegraphics[width=0.3\textwidth]{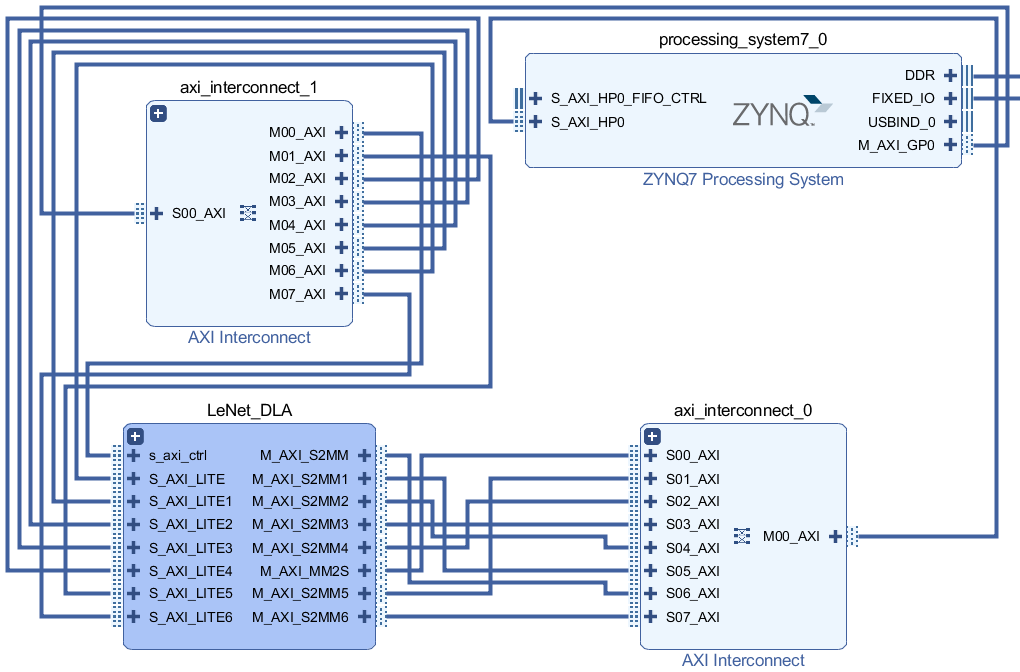}
\caption{The IP integration with the Zynq Processor}
\vspace{0mm}
\label{ip}
\end{figure}

To generate an Overlay that will be exported on the PYNQ board for the LeNet DLA, the generated IP is imported to Vivado where each axis port defined in Vivado HLS is now declared as AXI4-Stream port on the IP. An example of this can be seen in Fig. \ref{dma} where the LeNet\_DLA IP has ports representing Blobs for each layer.  The AXI4-Streams are written to and read from via AXI DMA as shown in Fig. \ref{dma}. Each of these AXI DMAs needs to interact with Python Overlay APIs to write data and read data from the AXI DMA. These connections in Fig. \ref{dma} are collapsed into a hierarchy called ``LeNet\_DLA" shown in Fig. \ref{ip} which is called in the Python environment. As shown in Fig. \ref{ip} of the block design, the LeNet\_DLA transfers data to and from the ZYNQ processor via the axi\_interconnect\_0 and axi\_interconnect\_1 modules, respectively. When all the connections are routed, the connections in the block diagram are validated, synthesized, and implemented. After the implementation, a bitstream file and a .tcl are generated which are exported to the PYNQ FPGA board to create an Overlay to be called at the Python environment.



As shown in the Hardware Deployment and On-board Verification phase (Section D) in the Fig. \ref{fig:DL1}, the Python Overlay API is imported into the Jupyter Notebook that allows reading from and writing to the IP on the hardware of the PYNQ board via AXI DMA. In the Python environment, the Python Overlay library uses AXI DMA APIs to call the AXI DMAs created in the block diagram directly and allows the writing of an image vector as a stream to the IP for processing. After the execution, the output of each layer is written to their respective AXI DMA, which is written to the Python environment. These outputs are verified with the SPVF and written to a specified file (File\_HW).  The prediction is read from the output register specified in Vivado HLS.

To illustrate this co-verification process, Fig. \ref{sample} shows how the output of conv1 layer defined in Caffe is written in Vivado HLS with its number of respective outputs. In Vivado HLS, the IN\_DATA and OUT\_CONV1 are defined as AXI4-Stream that allows the actual ports for the input image to be streamed in by the IN\_DATA and the Blob to be streamed out by OUT\_CONV1 as shown in Fig. \ref{sample}. Importing the IP into Vivado block design shown in Fig. \ref{sample} shows that IN\_DATA and OUT\_CONV1 have their own ports to be connected to an AXI DMA. OUT\_CONV1 is written to the conv1\_dma (which is shown in Image 6 of the code snippet shown in Fig. \ref{sample}) at the Python environment. Buffers are created and assigned to their AXI DMA for the data to be passed to and from the AXI DMA. Once the IP is signaled through the Python environment to start, the AXI DMA returns its values back to the buffer in which this buffer can be viewed in Python environment.

\begin{table*}[p]
     \begin{center}
     \caption{Snippet of results from layer-by-layer output of LeNet DLA implementation using default float (32-bit) data type. First column shows Caffe output (Software), second column shows Vivado HLS (Design) and third column shows PYNQ FPGA (Hardware)output results.}
     \begin{tabular}{ p{2cm}  p{5cm}  p{5cm}  p{5cm}}
     \toprule
      Layer & Caffe Output & Vivado HLS Output & PYNQ FPGA Output  \\
     \cmidrule(l){1-1}\cmidrule(l){2-2}\cmidrule(l){3-3}\cmidrule(l){4-4}
     Data
     \raisebox{-\totalheight}{\includegraphics[width=0.1\textwidth, height=10mm]{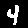}}\\
     \hline
     conv1
      &
     \raisebox{-\totalheight}{\includegraphics[width=0.26\textwidth, height=25mm]{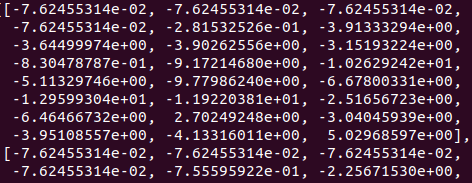}}
      &
     \raisebox{-\totalheight}{\includegraphics[width=0.26\textwidth, height=25mm]{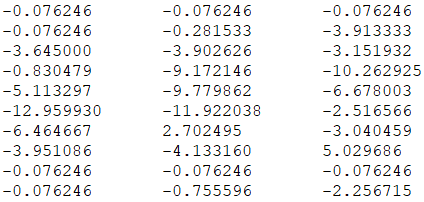}}
      &
     \raisebox{-\totalheight}{\includegraphics[width=0.26\textwidth, height=25mm]{Figs/conv1_V1.png}}\\
     \hline
     pool1
      &
     \raisebox{-\totalheight}{\includegraphics[width=0.26\textwidth, height=25mm]{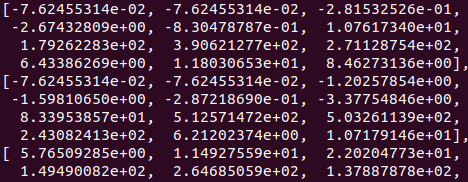}}
      &
     \raisebox{-\totalheight}{\includegraphics[width=0.26\textwidth, height=25mm]{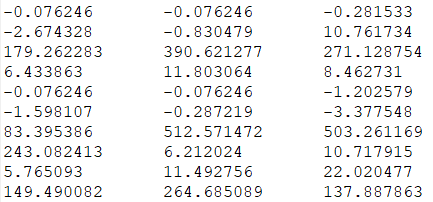}}
      &
     \raisebox{-\totalheight}{\includegraphics[width=0.26\textwidth, height=25mm]{Figs/pool1_V1.png}}\\
     \hline
     conv2
      &
     \raisebox{-\totalheight}{\includegraphics[width=0.26\textwidth, height=25mm]{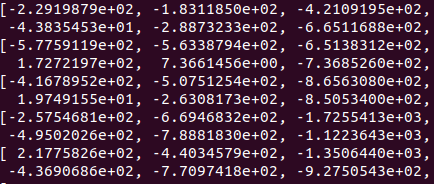}}
      &
     \raisebox{-\totalheight}{\includegraphics[width=0.26\textwidth, height=25mm]{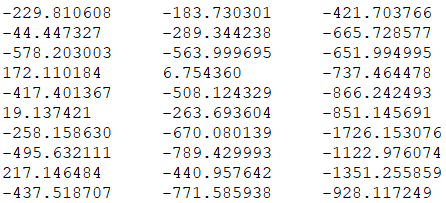}}
      &
     \raisebox{-\totalheight}{\includegraphics[width=0.26\textwidth, height=25mm]{Figs/conv2_V1.png}}\\
     \hline
     prediction
      &
     \raisebox{-\totalheight}{\includegraphics[width=0.2\textwidth, height=4mm]{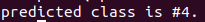}}
      &
     \raisebox{-\totalheight}{\includegraphics[width=0.2\textwidth, height=8mm]{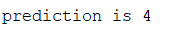}}
     &
     \raisebox{-\totalheight}{\includegraphics[width=0.2\textwidth, height=8mm]{Figs/prediction_V.png}}\\
     \hline
      \\ \bottomrule
      \end{tabular}
      \label{verification1}

      \caption{Snippet of results from layer-by-layer output of LeNet DLA implementation using Arbitrary Precision for bit-width reduction in hardware design and deployment. First column shows Caffe output (Software), shows Vivado HLS (Design) and third column shows PYNQ FPGA (Hardware) output results.}
     \begin{tabular}{ p{2cm}  p{5cm}  p{5cm}  p{5cm}}
     \toprule
      Layer & Caffe Output & Vivado HLS Output & PYNQ FPGA Output   \\
     \cmidrule(r){1-1}\cmidrule(lr){2-2}\cmidrule(l){3-3}\cmidrule(l){4-4}
     Data
     \raisebox{-\totalheight}{\includegraphics[width=0.1\textwidth, height=10mm]{Figs/A9.jpg}}\\
     \hline
     conv1
      &
     \raisebox{-\totalheight}{\includegraphics[width=0.26\textwidth, height=25mm]{Figs/conv1_C1.png}}
      &
     \raisebox{-\totalheight}{\includegraphics[width=0.26\textwidth, height=25mm]{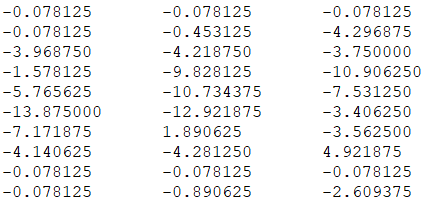}}
     &
     \raisebox{-\totalheight}{\includegraphics[width=0.26\textwidth, height=25mm]{Figs/conv1_P1.png}}\\
     \hline
     pool1
      &
     \raisebox{-\totalheight}{\includegraphics[width=0.26\textwidth, height=25mm]{Figs/pool1_C1.png}}
      &
     \raisebox{-\totalheight}{\includegraphics[width=0.26\textwidth, height=25mm]{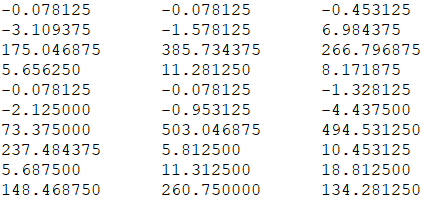}}
     &
     \raisebox{-\totalheight}{\includegraphics[width=0.26\textwidth, height=25mm]{Figs/pool1_P1.png}}\\
     \hline
     conv2
      &
     \raisebox{-\totalheight}{\includegraphics[width=0.26\textwidth, height=25mm]{Figs/conv2_C1.png}}
      &
     \raisebox{-\totalheight}{\includegraphics[width=0.26\textwidth, height=25mm]{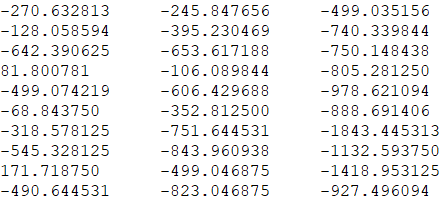}}
     &
     \raisebox{-\totalheight}{\includegraphics[width=0.26\textwidth, height=25mm]{Figs/conv2_P1.png}}\\
     \hline
     prediction
      &
     \raisebox{-\totalheight}{\includegraphics[width=0.2\textwidth, height=4mm]{Figs/ff.png}}
      &
     \raisebox{-\totalheight}{\includegraphics[width=0.2\textwidth, height=8mm]{Figs/prediction_V.png}}
     &
     \raisebox{-\totalheight}{\includegraphics[width=0.2\textwidth, height=8mm]{Figs/prediction_V.png}}\\
     \hline
      \\ \bottomrule
      \end{tabular}
      \label{verification2}
      \end{center}
      \end{table*}


      \begin{table*}[!h]
\begin{center}
\caption{Snippet of results from layer-by-layer output of Cifar-10 DLA implementation using default float (32-bit) data type. First column shows Caffe output (Software), second column shows Vivado HLS (Design) and third column shows PYNQ FPGA (Hardware)output results.}
     \begin{tabular}{ p{2cm}  p{5cm}  p{5cm}  p{5cm}}
     \toprule
      Layer & Caffe Output & Vivado HLS Output & PYNQ FPGA Output\\
     \cmidrule(r){1-1}\cmidrule(lr){2-2}\cmidrule(l){3-3}\cmidrule(l){4-4}
     Data
     \raisebox{-\totalheight}{\includegraphics[width=0.1\textwidth, height=15mm]{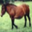}}\\
     \hline
     conv1
      &
     \raisebox{-\totalheight}{\includegraphics[width=0.28\textwidth, height=25mm]{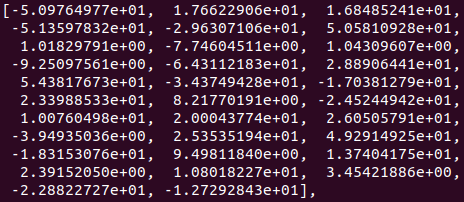}}
      &
     \raisebox{-\totalheight}{\includegraphics[width=0.27\textwidth, height=25mm]{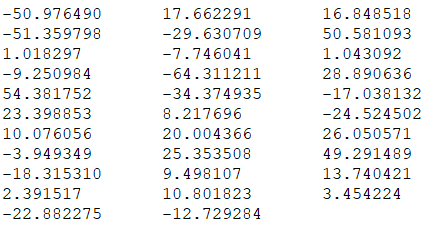}}
      &
     \raisebox{-\totalheight}{\includegraphics[width=0.27\textwidth, height=25mm]{Figs/cifar_viv_conv1.png}}\\
     \hline
     pool1
      &
     \raisebox{-\totalheight}{\includegraphics[width=0.28\textwidth, height=15mm]{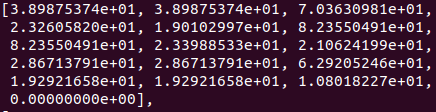}}
      &
     \raisebox{-\totalheight}{\includegraphics[width=0.27\textwidth, height=15mm]{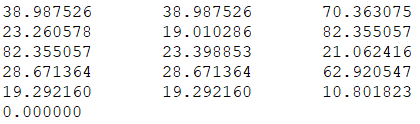}}
     &
     \raisebox{-\totalheight}{\includegraphics[width=0.27\textwidth, height=15mm]{Figs/cifar_viv_pool1.png}}\\
     \hline
     conv2
      &
     \raisebox{-\totalheight}{\includegraphics[width=0.27\textwidth, height=15mm]{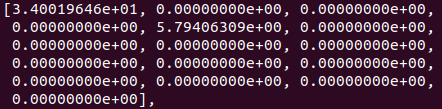}}
      &
     \raisebox{-\totalheight}{\includegraphics[width=0.27\textwidth, height=15mm]{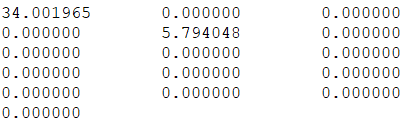}}
      &
     \raisebox{-\totalheight}{\includegraphics[width=0.27\textwidth, height=15mm]{Figs/cifar_viv_conv2.png}}\\
     \hline
     prediction
      &
     \raisebox{-\totalheight}{\includegraphics[width=0.15\textwidth, height=3mm]{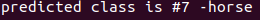}}
      &
     \raisebox{-\totalheight}{\includegraphics[width=0.15\textwidth, height=4mm]{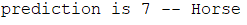}}
     &
     \raisebox{-\totalheight}{\includegraphics[width=0.15\textwidth, height=4mm]{Figs/cifar_viv_pred.png}}\\
     \hline
      \\ \bottomrule
      \end{tabular}
      \label{verification3}

      \caption{Snippet of results from layer-by-layer output of Cifar-10 DLA implementation using Arbitrary Precision for bit-width reduction in hardware design and deployment. First column shows Caffe output (Software), second column shows Vivado HLS (Design) and third column shows PYNQ FPGA (Hardware)output results.}
     \begin{tabular}{ p{2cm}  p{5cm}  p{5cm}  p{5cm}}
     \toprule
      Layer & Caffe Output & Vivado HLS & PYNQ FPGA Output  \\
     \cmidrule(r){1-1}\cmidrule(lr){2-2}\cmidrule(l){3-3}\cmidrule(l){4-4}
     Data
     \raisebox{-\totalheight}{\includegraphics[width=0.1\textwidth, height=15mm]{Figs/8.png}}\\
     \hline
     conv1
      &
     \raisebox{-\totalheight}{\includegraphics[width=0.28\textwidth, height=25mm]{Figs/cifar_conv1.png}}
      &
     \raisebox{-\totalheight}{\includegraphics[width=0.27\textwidth, height=25mm]{Figs/cifar_viv_conv1.png}}
      &
     \raisebox{-\totalheight}{\includegraphics[width=0.27\textwidth, height=25mm]{Figs/cifar_viv_conv1.png}}\\
     \hline
     pool1
      &
     \raisebox{-\totalheight}{\includegraphics[width=0.28\textwidth, height=15mm]{Figs/cifar_pool1.png}}
      &
     \raisebox{-\totalheight}{\includegraphics[width=0.27\textwidth, height=15mm]{Figs/cifar_viv_pool1.png}}
      &
     \raisebox{-\totalheight}{\includegraphics[width=0.27\textwidth, height=15mm]{Figs/cifar_viv_pool1.png}}\\
     \hline
     conv2
      &
     \raisebox{-\totalheight}{\includegraphics[width=0.27\textwidth, height=15mm]{Figs/cifar_conv2.png}}
      &
     \raisebox{-\totalheight}{\includegraphics[width=0.27\textwidth, height=15mm]{Figs/cifar_viv_conv2.png}}
           &
     \raisebox{-\totalheight}{\includegraphics[width=0.27\textwidth, height=15mm]{Figs/cifar_viv_conv2.png}}\\
     \hline
     prediction
      &
     \raisebox{-\totalheight}{\includegraphics[width=0.15\textwidth, height=3mm]{Figs/cifar_pred.png}}
      &
     \raisebox{-\totalheight}{\includegraphics[width=0.15\textwidth, height=4mm]{Figs/cifar_viv_pred.png}}
      &
     \raisebox{-\totalheight}{\includegraphics[width=0.15\textwidth, height=4mm]{Figs/cifar_viv_pred.png}}\\

     \hline
      \\ \bottomrule
      \end{tabular}
      \label{verification4}
      \end{center}
      \end{table*}

 Caffe software framework generated File\_SW at the end of Section A of Fig. \ref{fig:DL1}. The Vivado design simulation generated the layer-by-layer output feature of the DLA which is stored in File\_Design shown in Section D of Fig. \ref{fig:DL1}. Finally, the layer-by-layer output of the AXI DMA of each respective layer is written to File\_HW as depicted in Section E of Fig. \ref{fig:DL1}. Finally, as shown in Fig. \ref{fig:DL1}, the Section E of our 2L-3W co-verification compares the output of each layer at each stage of hardware-software co-design.

\section{Results and Lessons Learned}
\label{results}
The LeNet DLA for MNIST dataset and Caffe Cifar-10 inspired DLA for Cifar-10 datasets are shown in Figs. \ref{dma} and \ref{dma2}. They are implemented on the PYNQ hardware using the methodology shown in Fig. 2.

The LeNet DLA consists of 8 layers excluding the data and prob layers as shown in Fig. \ref{dma}. The data layer passes a 28x28 hand-written image of a digit through the layers designed in Caffe and also through the layers designed in Vivado HLS and the PYNQ FPGA. The results are shown in Table I.

\begin{table*}[t]
\caption{Table of Results of Similarity Scores for LeNet DLA}
\subfloat[Table of Similarity Scores and Parameters Compared for Layer-by-Layer Output of LeNet DLA When Hardware is Designed With Float Data Type]{
\begin{tabular}{|c|c|c|c|}
\hline
Layers                                       &              & Similarity Score      & Parameters Compared   \\ \hline
\multirow{4}{*}{conv1}                       & File\_SW     & \multirow{2}{*}{0.99999} & \multirow{4}{*}{3456} \\ \cline{2-2}
                                             & File\_Design &                       &                       \\ \cline{2-3}
                                             & File\_SW     & \multirow{2}{*}{0.99999} &                       \\ \cline{2-2}
                                             & File\_HW     &                       &                       \\ \hline
\multirow{4}{*}{pool1}                       & File\_SW     & \multirow{2}{*}{0.99999} & \multirow{4}{*}{864} \\ \cline{2-2}
                                             & File\_Design &                       &                       \\ \cline{2-3}
                                             & File\_SW     & \multirow{2}{*}{0.99999} &                       \\ \cline{2-2}
                                             & File\_HW     &                       &                       \\ \hline
\multicolumn{1}{|l|}{\multirow{4}{*}{conv2}} & File\_SW     & \multirow{2}{*}{0.98153} & \multirow{4}{*}{1024} \\ \cline{2-2}
\multicolumn{1}{|l|}{}                       & File\_Design &                       &                       \\ \cline{2-3}
\multicolumn{1}{|l|}{}                       & File\_SW     & \multirow{2}{*}{0.98153} &                       \\ \cline{2-2}
\multicolumn{1}{|l|}{}                       & File\_HW     &                       &                       \\ \hline
\multirow{4}{*}{pool2}                       & File\_SW     & \multirow{2}{*}{0.96887} & \multirow{4}{*}{256} \\ \cline{2-2}
                                             & File\_Design &                       &                       \\ \cline{2-3}
                                             & File\_SW     & \multirow{2}{*}{0.96887} &                       \\ \cline{2-2}
                                             & File\_HW     &                       &                       \\ \hline
\multirow{4}{*}{conv3}                       & File\_SW     & \multirow{2}{*}{0.99057} & \multirow{4}{*}{120} \\ \cline{2-2}
                                             & File\_Design &                       &                       \\ \cline{2-3}
                                             & File\_SW     & \multirow{2}{*}{0.99057} &                       \\ \cline{2-2}
                                             & File\_HW     &                       &                       \\ \hline
\multirow{4}{*}{fc1}                         & File\_SW     & \multirow{2}{*}{0.99333} & \multirow{4}{*}{84}  \\ \cline{2-2}
                                             & File\_Design &                       &                       \\ \cline{2-3}
                                             & File\_SW     & \multirow{2}{*}{0.99333} &                       \\ \cline{2-2}
                                             & File\_HW     &                       &                       \\ \hline
\multirow{4}{*}{fc2}                         & File\_SW     & \multirow{2}{*}{0.99088} & \multirow{4}{*}{10} \\ \cline{2-2}
                                             & File\_Design &                       &                       \\ \cline{2-3}
                                             & File\_SW     & \multirow{2}{*}{0.99088} &                       \\ \cline{2-2}
                                             & File\_HW     &                       &                       \\ \hline
\label{aplenet}
\end{tabular}}
\qquad
\subfloat[Table of Similarity Scores and Parameters Compared for Layer-by-Layer Output of LeNet DLA When Hardware is Designed With Arbitrary Precision Data Type]{
\begin{tabular}{|c|c|c|c|}
\hline
Layers                                       &              & Similarity Score      & Parameters Compared   \\ \hline
\multirow{4}{*}{conv1}                       & File\_SW     & \multirow{2}{*}{0.82564} & \multirow{4}{*}{3456} \\ \cline{2-2}
                                             & File\_Design &                       &                       \\ \cline{2-3}
                                             & File\_SW     & \multirow{2}{*}{0.82564} &                       \\ \cline{2-2}
                                             & File\_HW     &                       &                       \\ \hline
\multirow{4}{*}{pool1}                       & File\_SW     & \multirow{2}{*}{0.84104} & \multirow{4}{*}{864} \\ \cline{2-2}
                                             & File\_Design &                       &                       \\ \cline{2-3}
                                             & File\_SW     & \multirow{2}{*}{0.84104} &                       \\ \cline{2-2}
                                             & File\_HW     &                       &                       \\ \hline
\multicolumn{1}{|l|}{\multirow{4}{*}{conv2}} & File\_SW     & \multirow{2}{*}{0.74304} & \multirow{4}{*}{1024} \\ \cline{2-2}
\multicolumn{1}{|l|}{}                       & File\_Design &                       &                       \\ \cline{2-3}
\multicolumn{1}{|l|}{}                       & File\_SW     & \multirow{2}{*}{0.74304} &                       \\ \cline{2-2}
\multicolumn{1}{|l|}{}                       & File\_HW     &                       &                       \\ \hline
\multirow{4}{*}{pool2}                       & File\_SW     & \multirow{2}{*}{0.68196} & \multirow{4}{*}{256} \\ \cline{2-2}
                                             & File\_Design &                       &                       \\ \cline{2-3}
                                             & File\_SW     & \multirow{2}{*}{0.68196} &                       \\ \cline{2-2}
                                             & File\_HW     &                       &                       \\ \hline
\multirow{4}{*}{conv3}                       & File\_SW     & \multirow{2}{*}{0.64940} & \multirow{4}{*}{120} \\ \cline{2-2}
                                             & File\_Design &                       &                       \\ \cline{2-3}
                                             & File\_SW     & \multirow{2}{*}{0.64940} &                       \\ \cline{2-2}
                                             & File\_HW     &                       &                       \\ \hline
\multirow{4}{*}{fc1}                         & File\_SW     & \multirow{2}{*}{0.66700} & \multirow{4}{*}{84}  \\ \cline{2-2}
                                             & File\_Design &                       &                       \\ \cline{2-3}
                                             & File\_SW     & \multirow{2}{*}{0.66700} &                       \\ \cline{2-2}
                                             & File\_HW     &                       &                       \\ \hline
\multirow{4}{*}{fc2}                         & File\_SW     & \multirow{2}{*}{0.77909} & \multirow{4}{*}{10} \\ \cline{2-2}
                                             & File\_Design &                       &                       \\ \cline{2-3}
                                             & File\_SW     & \multirow{2}{*}{0.77909} &                       \\ \cline{2-2}
                                             & File\_HW     &                       &                       \\ \hline
\label{aplenet}
\end{tabular}}
\label{table_simi}
\end{table*}

\begin{table*}[t]
\caption{Table of Results of Similarity Scores for Cifar-10 DLA}
\subfloat[Table of Similarity Scores and Parameters Compared for Layer-by-Layer Output of Cifar-10 DLA When Hardware is Designed With Float Data Type]{
\begin{tabular}{|c|c|c|c|}
\hline
Layers                                       &              & Similarity Score      & Parameters Compared   \\ \hline
\multirow{4}{*}{conv1}                       & File\_SW     & \multirow{2}{*}{0.99889} & \multirow{4}{*}{5120} \\ \cline{2-2}
                                             & File\_Design &                       &                       \\ \cline{2-3}
                                             & File\_SW     & \multirow{2}{*}{0.99889} &                       \\ \cline{2-2}
                                             & File\_HW     &                       &                       \\ \hline
\multirow{4}{*}{pool1_relu1}                       & File\_SW     & \multirow{2}{*}{0.99885} & \multirow{4}{*}{1280} \\ \cline{2-2}
                                             & File\_Design &                       &                       \\ \cline{2-3}
                                             & File\_SW     & \multirow{2}{*}{0.99885} &                       \\ \cline{2-2}
                                             & File\_HW     &                       &                       \\ \hline
\multicolumn{1}{|l|}{\multirow{4}{*}{conv2_relu2}} & File\_SW     & \multirow{2}{*}{1.00510} & \multirow{4}{*}{2560} \\ \cline{2-2}
\multicolumn{1}{|l|}{}                       & File\_Design &                       &                       \\ \cline{2-3}
\multicolumn{1}{|l|}{}                       & File\_SW     & \multirow{2}{*}{1.00510} &                       \\ \cline{2-2}
\multicolumn{1}{|l|}{}                       & File\_HW     &                       &                       \\ \hline
\multirow{4}{*}{pool2}                       & File\_SW     & \multirow{2}{*}{0.99896} & \multirow{4}{*}{640} \\ \cline{2-2}
                                             & File\_Design &                       &                       \\ \cline{2-3}
                                             & File\_SW     & \multirow{2}{*}{0.99896} &                       \\ \cline{2-2}
                                             & File\_HW     &                       &                       \\ \hline
\multirow{4}{*}{conv3_relu3}                       & File\_SW     & \multirow{2}{*}{0.99732} & \multirow{4}{*}{960} \\ \cline{2-2}
                                             & File\_Design &                       &                       \\ \cline{2-3}
                                             & File\_SW     & \multirow{2}{*}{0.99732} &                       \\ \cline{2-2}
                                             & File\_HW     &                       &                       \\ \hline
\multirow{4}{*}{pool3}                         & File\_SW     & \multirow{2}{*}{1.00856} & \multirow{4}{*}{240}  \\ \cline{2-2}
                                             & File\_Design &                       &                       \\ \cline{2-3}
                                             & File\_SW     & \multirow{2}{*}{1.00856} &                       \\ \cline{2-2}
                                             & File\_HW     &                       &                       \\ \hline
\multirow{4}{*}{fc1}                         & File\_SW     & \multirow{2}{*}{0.98541} & \multirow{4}{*}{50} \\ \cline{2-2}
                                             & File\_Design &                       &                       \\ \cline{2-3}
                                             & File\_SW     & \multirow{2}{*}{0.98541} &                       \\ \cline{2-2}
                                             & File\_HW     &                       &                       \\ \hline
\multirow{4}{*}{fc2}                         & File\_SW     & \multirow{2}{*}{0.99964} & \multirow{4}{*}{10} \\ \cline{2-2}
                                             & File\_Design &                       &                       \\ \cline{2-3}
                                             & File\_SW     & \multirow{2}{*}{0.99964} &                       \\ \cline{2-2}
                                             & File\_HW     &                       &                       \\ \hline
\label{cifarfloat}
\end{tabular}}
\qquad
\subfloat[Table of Similarity Scores and Parameters Compared for Layer-by-Layer Output of Cifar-10 DLA When Hardware is Designed With Arbitrary Precision Data Type]{
\begin{tabular}{|c|c|c|c|}
\hline
Layers                                       &              & Similarity Score      & Parameters Compared   \\ \hline
\multirow{4}{*}{conv1}                       & File\_SW     & \multirow{2}{*}{0.99889} & \multirow{4}{*}{5120} \\ \cline{2-2}
                                             & File\_Design &                       &                       \\ \cline{2-3}
                                             & File\_SW     & \multirow{2}{*}{0.99889} &                       \\ \cline{2-2}
                                             & File\_HW     &                       &                       \\ \hline
\multirow{4}{*}{pool1_relu1}                       & File\_SW     & \multirow{2}{*}{0.99885} & \multirow{4}{*}{1280} \\ \cline{2-2}
                                             & File\_Design &                       &                       \\ \cline{2-3}
                                             & File\_SW     & \multirow{2}{*}{0.99885} &                       \\ \cline{2-2}
                                             & File\_HW     &                       &                       \\ \hline
\multicolumn{1}{|l|}{\multirow{4}{*}{conv2_relu2}} & File\_SW     & \multirow{2}{*}{1.00510} & \multirow{4}{*}{2560} \\ \cline{2-2}
\multicolumn{1}{|l|}{}                       & File\_Design &                       &                       \\ \cline{2-3}
\multicolumn{1}{|l|}{}                       & File\_SW     & \multirow{2}{*}{1.00510} &                       \\ \cline{2-2}
\multicolumn{1}{|l|}{}                       & File\_HW     &                       &                       \\ \hline
\multirow{4}{*}{pool2}                       & File\_SW     & \multirow{2}{*}{0.99896} & \multirow{4}{*}{640} \\ \cline{2-2}
                                             & File\_Design &                       &                       \\ \cline{2-3}
                                             & File\_SW     & \multirow{2}{*}{0.99896} &                       \\ \cline{2-2}
                                             & File\_HW     &                       &                       \\ \hline
\multirow{4}{*}{conv3_relu3}                       & File\_SW     & \multirow{2}{*}{0.99732} & \multirow{4}{*}{960} \\ \cline{2-2}
                                             & File\_Design &                       &                       \\ \cline{2-3}
                                             & File\_SW     & \multirow{2}{*}{0.99732} &                       \\ \cline{2-2}
                                             & File\_HW     &                       &                       \\ \hline
\multirow{4}{*}{pool3}                         & File\_SW     & \multirow{2}{*}{1.00856} & \multirow{4}{*}{240}  \\ \cline{2-2}
                                             & File\_Design &                       &                       \\ \cline{2-3}
                                             & File\_SW     & \multirow{2}{*}{1.00856} &                       \\ \cline{2-2}
                                             & File\_HW     &                       &                       \\ \hline
\multirow{4}{*}{fc1}                         & File\_SW     & \multirow{2}{*}{0.98541} & \multirow{4}{*}{50} \\ \cline{2-2}
                                             & File\_Design &                       &                       \\ \cline{2-3}
                                             & File\_SW     & \multirow{2}{*}{0.98541} &                       \\ \cline{2-2}
                                             & File\_HW     &                       &                       \\ \hline
\multirow{4}{*}{fc2}                         & File\_SW     & \multirow{2}{*}{0.99964} & \multirow{4}{*}{10} \\ \cline{2-2}
                                             & File\_Design &                       &                       \\ \cline{2-3}
                                             & File\_SW     & \multirow{2}{*}{0.99964} &                       \\ \cline{2-2}                                           & File\_HW     &                       &                       \\ \hline
\label{cifarap}
\end{tabular}}
\label{table_simi}
\end{table*}

The Tables \ref{verification1} and \ref{verification3} show the values of subsections of the array outputted by the Conv1, Pool1 and Conv2 layers of these respective written files of LeNet and DLA for Cifar-10 DLA respectively. The 3-way verification performed by the Python script which compares of the output values of each layer and the prediction written to the files returns a similarity score per layer. The similarity score is defined as the metric for
measuring element-by-element similarity in terms of magnitude of the values stored in the arrays produced by each layer and written to the three files (File SW, File Design, File HW).

The similarity score per layer for the design stage ($SC_{Des}$) is given as:
\begin{equation}
\centering
SC_{Des} = \dfrac{\sum_{i=0}^n ( 1 - \dfrac{X_{1} - Y_{1}}{X_{1}})}{n}
\end{equation}

$where: $\\
$SC_{Des} = $Similarity score for a layer in design stage\\
$i = $$ ith$ element written to a particular file\\
$n = $Number of parameters to be compared in the layer\\
$X_{1} = Max(|E_i|_{SW} , |E_i|_{Des})$ \\
$Y_{1} = Min(|E_i|_{SW} , |E_i|_{Des})$\\
$|E_i|_{SW} = $Absolute value of the $ith$ element value written to the File\_SW file\\
$|E_i|_{Des} = $Absolute value of the $ith$ corresponding element value written to the File\_Design file\\

Similarly, the similarity score per layer for the deployment stage ($SC_{HW}$) is given as:
\begin{equation}
\centering
SC_{HW} = \dfrac{\sum_{i=0}^n ( 1 - \dfrac{X_{2} - Y_{2}}{X_{2}})}{n}
\end{equation}

$where: $\\
$SC_{HW} = $Similarity score for a layer in hardware deployment stage\\
$i = $$ ith$ element written to a particular file\\
$n = $Number of parameters to be compared in the layer\\
$X_{2} = Max(|E_i|_{SW} , |E_i|_{HW})$ \\
$Y_{2} = Min(|E_i|_{SW} , |E_i|_{HW})$\\
$|E_i|_{SW} = $Absolute value of the $ith$ element value written to the File\_SW file\\
$|E_i|_{HW} = $Absolute value of the $ith$ element value written to the File\_HW file\\
$|E_i|_{Des} = $Absolute value of the $ith$ corresponding element value written to the File\_Design file\\

Table \ref{verification1} shows snippets of partial results of the layer-by-layer output values written to the File\_SW file in the software stage, the File\_Design file in the design stage of the LeNet DLA and the File\_HW in the hardware deployment stage to obtain similarity scores in the design stage and deployment stage respectively. 

Prior to the design stage, the training of the DLA is done in Caffe software environment using float (32-bits precision) data type. Hence the parameters and the Blobs of the DLA are in float data type. The layer-by-layer output (Blob) are obtained and written to File\_SW. In the design stage, the DLA is simulated with parameters and Blobs of float data type numbers to obtain and write the layer-by-layer output in the design stage to File\_Design. The values written to the File\_Design are verified and compared with the layer-by-layer outputs written to File\_SW to obtain the similarity scores at the design stage as shown in Table \ref{cifarfloat}. Once the result shows desirable similarity scores, an IP is generated from the hardware design and exported and configured in Vivado to generate a bit-stream file that is deployed on the PYNQ FPGA board. The layer-by-layer values outputted by the PYNQ FPGA after deployment are obtained are written to File\_HW to obtain the layer-by-layer similarity score for the deployment stage. The similarity score for the deployment stage for LeNet DLA is shown in Table \ref{cifarfloat}.

From Table \ref{cifarfloat}, a 99\% similarity score for each layer is obtained for the LeNet DLA in the design stage and deployment stage using float data type.

FPGAs have a common characteristic of having limited area and hardware resources (DSPs, LUTs, Flip-flops, BRAM). For scalability, one of the strategies to ensure the large DLA fit the FPGA boards, the bit-width of parameters and Blob precisions of the large DLA are truncated using Arbitrary Precision (AP) libraries provided in the hardware design stage in Vivado HLS. This truncation reduces the memory and computation requirement of the large DLA. For the LeNet DLA in this work, the parameters and Blobs are truncated from 32-bit precision to 8-bits and 24-bits precisions respectively. The truncation reduces the area of the DLA on the board without compromising on accuracy as truncated parameters and Blobs are tested with 100 images and they show consistent predictions with the hardware design and deployment using float data type. This truncation leads to changes in the values of the parameters and hence changes in Blobs as shown in Table \ref{verification2}. The similarity score of the design and deployment stage are obtained as shown in Table \ref{aplenet}. 

From Table \ref{aplenet}, similarity scores ranging from 65\% - 84\% is obtained in the design and deployment stage when layer-by-layer output values written in File\_Design and File\_HW are compared with layer-by-layer output values written to File\_SW. This drop in similarity scores is due to bit-width truncation of the parameters and Blobs of the LeNet DLA in the design stage and hence the deployment stage. 

The 3-way prediction values written to the files are equivalent and consistent. This depicts the successful implementation of a DLA on the PYNQ FPGA. Based on the similarity score provided by the Python script, recommendations can be made on where to debug or redesign if the similarity score is below a certain threshold. This helps to avoid blind debugging of results during hardware implementation of DLA. A total of 100 images are used to validate our 3-way verification methodology and it turns out to be consistent in all cases.

Tables \ref{verification3} and \ref{verification4} shows a similar result of the implementation of a DLA shown in Fig. \ref{dma2} for Cifar-10 dataset. The DLA also consist of 8 layers and accepts a 32x32 input image. Just like Tables \ref{verification1} and \ref{verification2},  Tables \ref{verification3} and \ref{verification4} also shows subsections of the 2D-matrices that are written to the File\_SW, File\_Design and File\_HW for float data type and implementation using arbitrary precision data type for bit-width reduction respectively. The Python script returns a 3-way similarity score of 99\% for the values written to the files in the design stage(File\_Design) and deployment stage (File\_HW) using float data type when compared with File\_SW as seen in Table \ref{cifarfloat} and a similarity score ranging from 65\% - 84\% as seen in \ref{cifarap} for the values written to the files in the design stage (File\_Design) and deployment stage (File\_HW) using arbitrary precision data type when compared with File\_SW. The arbitrary precision prediction show consistent results with the prediction obtained using float data type for 100 images.

\section{Related Work}
Several approaches in existing literature have been adopted to achieve efficient mapping of DLAs to FPGA boards. Guo et. al \cite{guo2017angel} proposes a design flow for mapping CNNs onto embedded FPGA. In \cite{guo2017angel}, data quantization is introduced to reduce the bit-width of CNN models to achieve smaller memory and computation requirements with negligible accuracy loss. A compiler that maps the CNN to the FPGA is also proposed.

Florian et. al \cite{kastner2018hardware} proposes a tool flow for the hardware/software codesign implementation of CNNs on PYNQ FPGAs. FPGA possess Dynamic Partial Reconfiguration (DPR) capabilities that enable the exchange of logic partitions within the FPGA fabric. This property offers a major advantage for designing hardware architectures able to adapt and reconfigure the hardware due to characteristics of DLA using high-level synthesis.

Jiandong et. al \cite{mu2018collaborative} proposes a collaborative framework to optimize the OpenCL based CNN design for CNN applications. The introduction of LoopTree to capture the main features of OpenCL based hardware design.  Hardware design specifications like loop orders, loop tiling, Block RAMs (BRAM) and  Double Data Rate (DDR) configurations, and OpenCL attributes are utilized. Then a coarse-grained model is employed in evaluating the performance of LoopTree and to find candidate designs. Finally, a fine-grained model is employed to tune the candidate designs to obtain the best design deployed on the hardware. Also, \cite{han2016eie}, proposes weight compression and weight sharing neural networks in order to allow for the proper hardware resource utilization that enables the large neural network models to fit in ASICs and FPGAs.

 Xiang et. al  \cite{xiang2018output} proposes a software simulation-based approach for the verification of Multilayer Neural Networks by coming up with an algorithm to measure the maximum sensitivity for the output of a finite number of different simulations corresponding to different finite bounded inputs. The sensitivity of the network is given as the mathematical expectation of output deviations due to input and weight deviations with respect to overall input and weight values in a given continuous interval. The maximum sensitivity used to measure the maximum deviation of outputs, which is brought by bounded disturbances around the input. The maximum sensitivity represents the output reachable sets of the network and is measured and computed layer-by-layer. These measurements are used for the verification of the layer-by-layer output of the network.

Dwarakanath et. al \cite{dwarakanath2018identifying} proposes a software-based approach of verification of machine learning-based image classifiers using metamorphic testing. This approach builds multiple relationships between the subsequent output of a classifier to different inputs to derive the degree of correctness of the implementation of the classifier. This approach is designed to detect implementation bugs in the implementation of the classifier. The metamorphic testing comes up with different permutations of cases for the training and testing input features, training instances and layers and also scaling of the test data samples of the image classifier to generate different outputs.

Choi et. al \cite{choi2018stochastic} proposes a stochastic functional verification method in designing DNN-based systems. In this approach, synthetic data sets are generated in a virtual environment and added to the training set for a DNN. The DNN is trained with both dataset and validated with validation subsets of both datasets. A comparison metric such as class-wise average precision is used to compare the performance of the model on both validation datasets against a predefined threshold.  For a DNN under verification, the DNN is trained with synthetic datasets and the comparison, metric is obtained. The similarity between the comparison metric and the predefined threshold is used to validate the verification.

Cong et. al \cite{hao2019fpga} proposes a time saving co-design methodology that simultaneously searches possible design options to auto-generate efficient DNNs optimized for FPGA deployment. \cite{hao2019fpga} introduces a template for the generation of DNN with efficient performance and hardware resource utilization. An automatic HLS generator is proposed to help translate the auto-generated DNN to synthesizable C code for hardware deployment.

In reference \cite{changwoolee} is a Github repository that the C++ code (Design code) for mapping LeNet DLA on hardware. The repository shows the weights and algorithms of each layer. This code is an already finished DLA on an FPGA board. This repository does not give information about which framework has been used to train the DLA and does not provide a means of debugging and validating the output of each layer in order to accomplish design time verification at every stage.

The references \cite{stephendigikey} and \cite{alveo} shows an introduction to the deployment of Machine Learning on Hardware. This only shows stacks and block diagrams of how neural networks is utilized on hardware and also the number of parameters and MACC (Multiply-Accumulate) units required by a DLA. This does not give a full picture from the training to the testing and successful deployment of DLAs on FPGA boards and other hardware.

These approaches are either limited to the software environment or they do not take into consideration the verification of the implementation correctness of the DLA mapping onto hardware across all the design stages involved.

\begin{table*}[t]
\center
  \caption{Verification Approach Comparison With Other Works.}
 \vspace{0mm}
\begin{tabular}{|c|c|c|c|c|c|c|c|c|c|}
\hline
\multicolumn{1}{|l|}{}                                                   & \multicolumn{1}{c|}{Approach} & \multicolumn{1}{l|}{{\cite{guo2017angel}}} & \multicolumn{1}{l|}{{\cite{kastner2018hardware}}} & \multicolumn{1}{l|}{{\cite{mu2018collaborative}}} & \multicolumn{1}{l|}{{\cite{xiang2018output}}} & \multicolumn{1}{l|}{{\cite{dwarakanath2018identifying}}} & \multicolumn{1}{l|}{{\cite{choi2018stochastic}}} & \multicolumn{1}{l|}{{\cite{hao2019fpga}}} & 2L-3W\\ \hline
\multirow{2}{*}{Software Verification}                                   & Layer-by-Layer Verification   & x                            & x                            & x                            & $\surd$                             & $\surd$                             & $\surd$                            & x                            & $\surd$ \\ \cline{2-10}
                                                                         & Final Layer Verification      & $\surd$                             & $\surd$                            & x                            & $\surd$                             & $\surd$                             & $\surd$                             & $\surd$                             & $\surd$ \\ \hline
\multirow{2}{*}{Hardware Verification}                                   & Layer-by-Layer Verification   & x                            & x                            & x                            & x                            & x                            & x                            & x                            & $\surd$ \\ \cline{2-10}
                                                                         & Final Layer Verification      & $\surd$                             & $\surd$                            & $\surd$                             & x                            & x                            & x                            & $\surd$                             & $\surd$ \\ \hline
\multicolumn{1}{|l|}{\multirow{2}{*}{Hardware-Software Co-Verification}} & Layer-by-Layer Verification   & x                            & x                            & x                            & x                            & x                            & x                            & x                            & $\surd$ \\ \cline{2-10}
\multicolumn{1}{|l|}{}                                                   & Final Layer Verification      & x                            & x                            & x                            & x                            & x                            & x                            & x                            & $\surd$ \\ \hline
\end{tabular}
\label{tab:compare}
\end{table*}

\section{Comparison With State-of-the-Art}
Some state-of-the-art approaches have been adopted to ascertain the implementation correctness of DLA.
Guo et. al \cite{guo2017angel} proposes an approach that allows for the hardware-software co-design of DLA on FPGAs. This approach only has a means of validating the DLA at the final layers of the software and hardware. The limitation of this approach is that it does not account for layer-by-layer verification of the output of the layers.

Florian et. al \cite{kastner2018hardware} proposes a toolflow approach for the hardware-software co-design of DLA on FPGAs. The means of validating the co-design is at the final layers of the software and hardware. This toolflow approach does not take into consideration the verification of the layer-by-layer outputs to ensure the implementation correctness on the hardware. The approach also does not provide a means of debugging in case of errors.

Jiandong et. al \cite{mu2018collaborative} proposes a collaborative framework to optimize the deployment of DLA on FPGA. This approach validates the correctness of the deployment of the DLA only at the final layers of the hardware deployment. The limitation of this approach is that it does not account for software implementation, and at the hardware level, it does not provide layer-by-layer verification of the DLA.

Xiang et. al \cite{xiang2018output} proposes a software simulation based approach to verify the correctness of a DLA. The verification approach is limited to the layer-by-layer output and the accuracy of the final prediction. This approach does not provide a means of verifying the implementation correctness of the mapping of DLAs on hardware.

Dwarakanath et.al \cite{dwarakanath2018identifying} proposes a software based approach to verify the correctness of a image classifiers. The verification approach verifies to the layer-by-layer output and the accuracy of the final prediction only. This approach does not take into consideration an approach that can be applied to the mapping of DLAs to FPGA boards.

Choi et. al \cite{choi2018stochastic} introduces a stochastic functional verification method using synthetic datasets. This method verifies the layer-by-layer output and the accuracy of the deep learning model. This approach is not scalable when trying to achieve successful mapping of DLAs on FPGA boards.

Cong et. al \cite{hao2019fpga} proposes a co-design methodology that simultaneously generates a software design model and an synthesizable C code for the hardware design. This approach only validates the design based on the accuracy of prediction of the model.

Table \ref{tab:compare} shows that our proposed method can verify all the six types of cross layer verification.


\section{Conclusions}
\label{conclusion}

This work proposes a 2-Level 3-Way methodology for hardware-software co-verification of DLA from deep learning software framework to HLS design of DLA and finally onto DLA deployment on the FPGA board. This methodology is used to test the hardware implementation correctness of 2 DLAs (LeNet and Caffe inspired Cifar-10 network) on PYNQ FPGA board.  To the best of author's knowledge this is the first time a methodology is developed, which performs layer-by-layer co-verification for mapping of DLA architectures across the 3 paradigms (software, design and hardware level). The methodology can help to achieve successful implementation and mapping of DLA onto FPGA during the design phase and can help in the cross paradigm debugging process. We proposed a new metric for cross paradigm co-verification, called similarity score, which  as a metric to measure the degree of correctness of the implementation of each layer. The similarity score also helps to show layers that need debugging.  Our implementation results from Caffe software to Vivado HLS design and finally to Xilinx's PYNQ FPGA show similarity scores of 99\% for LeNet and Caffe inspired Cifar-10 network in the design stage. A range of similarity scores from 65\% - 84\% are obtained in the deployment stage due to truncation of the bit-width of the LeNet DLA so it can fit on the PYNQ FPGA board. This stipulates the successful mapping of the DLA onto the PYNQ FPGA board

\bibliographystyle{IEEEtran}
\bibliography{CC}	

\begin{thebibliography}{10}
\providecommand{\url}[1]{#1}
\csname url@samestyle\endcsname
\providecommand{\newblock}{\relax}
\providecommand{\bibinfo}[2]{#2}
\providecommand{\BIBentrySTDinterwordspacing}{\spaceskip=0pt\relax}
\providecommand{\BIBentryALTinterwordstretchfactor}{4}
\providecommand{\BIBentryALTinterwordspacing}{\spaceskip=\fontdimen2\font plus
\BIBentryALTinterwordstretchfactor\fontdimen3\font minus
  \fontdimen4\font\relax}
\providecommand{\BIBforeignlanguage}[2]{{%
\expandafter\ifx\csname l@#1\endcsname\relax
\typeout{** WARNING: IEEEtran.bst: No hyphenation pattern has been}%
\typeout{** loaded for the language `#1'. Using the pattern for}%
\typeout{** the default language instead.}%
\else
\language=\csname l@#1\endcsname
\fi
#2}}
\providecommand{\BIBdecl}{\relax}
\BIBdecl

\bibitem{zhang2015optimizing}
C.~Zhang, P.~Li, G.~Sun, Y.~Guan, B.~Xiao, and J.~Cong, ``{Optimizing
  FPGA-based accelerator design for deep convolutional neural networks},'' in
  \emph{Proceedings of the 2015 ACM/SIGDA International Symposium on
  Field-Programmable Gate Arrays}.\hskip 1em plus 0.5em minus 0.4em\relax ACM,
  2015, pp. 161--170.

\bibitem{tolu2}
T.~A. Odetola, O.~Ogheneuriri, and S.~R. Hasan, ``A scalable multilabel
  classification to deploy deep learning architectures for edge devices,''
  \emph{arXiv preprint arXiv:1911.02098}, 2019.

\bibitem{wang2017dlau}
C.~Wang, L.~Gong, Q.~Yu, X.~Li, Y.~Xie, and X.~Zhou, ``{DLAU: A scalable deep
  learning accelerator unit on FPGA},'' \emph{IEEE Transactions on
  Computer-Aided Design of Integrated Circuits and Systems}, vol.~36, no.~3,
  pp. 513--517, 2017.

\bibitem{baza2019b}
M.~Baza, N.~Lasla, M.~Mahmoud, and M.~Abdallah, ``B-ride: Ride sharing with
  privacy-preservation, trust and fair payment atop public blockchain,''
  \emph{arXiv preprint arXiv:1906.09968}, 2019.

\bibitem{baza2018blockchain}
M.~Baza, M.~Nabil, N.~Lasla, K.~Fidan, M.~Mahmoud, and M.~Abdallah,
  ``Blockchain-based firmware update scheme tailored for autonomous vehicles,''
  \emph{Proc. of the IEEE Wireless Communications and Networking Conference
  (WCNC), Marrakech, Morocco}, April 2019.

\bibitem{parksmarnet}
W.~Al~Amiri, M.~Baza, M.~Mahmoud, K.~Banawan, W.~Alasmary, and K.~Akkaya,
  ``Privacy-preserving smart parking system using blockchain and private
  information retrieval,'' \emph{Proc. of the IEEE International Conference on
  Smart Applications, Communications and Networking (SmartNets 2019)}, 2020.

\bibitem{baza2019blockchain}
M.~Baza, M.~Nabil, M.~Ismail, M.~Mahmoud, E.~Serpedin, and M.~Rahman,
  ``Blockchain-based charging coordination mechanism for smart grid energy
  storage units,'' \emph{Proc. Of IEEE International Conference on Blockchain,
  Atlanta, USA}, July, 2019.

\bibitem{parkccnc}
W.~Al~Amiri, M.~Baza, K.~Banawan, M.~Mahmoud, W.~Alasmary, and K.~Akkaya,
  ``Towards secure smart parking system using blockchain technology,''
  \emph{Proc. of 17th IEEE Annual Consumer Communications $\&$ Networking
  Conference (CCNC), Las vegas, USA}, 2020.

\bibitem{pazos2019privacy}
M.~Baza, M.~Pazos-Revilla, M.~Nabil, A.~Sherif, M.~Mahmoud, and W.~Alasmary,
  ``Privacy-preserving and collusion-resistant charging coordination schemes
  for smart grid,'' \emph{arXiv preprint arXiv:1905.04666}, 2019.

\bibitem{baza2019detecting}
M.~Baza, M.~Nabil, N.~Bewermeier, K.~Fidan, M.~Mahmoud, and M.~Abdallah,
  ``Detecting sybil attacks using proofs of work and location in vanets,''
  \emph{arXiv preprint arXiv:1904.05845}, 2019.

\bibitem{Lightride}
M.~Baza, M.~Mahmoud, G.~Srivastava, W.~Alasmary, and M.~Younis, ``A light
  blockchain-powered privacy-preserving organization scheme for ride sharing
  services,'' \emph{Proc. of the IEEE 91th Vehicular Technology Conference
  (VTC-Spring), Antwerp, Belgium}, May 2020.

\bibitem{Andrew}
M.~Baza, A.~Salazar, M.~Mahmoud, M.~Abdallah, and K.~Akkaya, ``On sharing
  models instead of data using mimic learning for smart health applications,''
  \emph{Proc. of the IEEE International Conference on Informatics, IoT, and
  Enabling Technologies (ICIoT'20) , Doha, Qatar}, Feb. 2020.

\bibitem{shafee2019mimic}
A.~Shafee, M.~Baza, D.~A. Talbert, M.~M. Fouda, M.~Nabil, and M.~Mahmoud,
  ``Mimic learning to generate a shareable network intrusion detection model,''
  \emph{Proc. of the IEEE Consumer Communications \& Networking Conference,Las
  Vegas, USA}, 2020.

\bibitem{baza2015efficient}
M.~Baza, M.~M. Fouda, A.~S.~T. Eldien, and H.~A. Mansour, ``An efficient
  distributed approach for key management in microgrids,'' \emph{Proc. of the
  Computer Engineering Conference (ICENCO), Egypt}, pp. 19--24, 2015.

\bibitem{blockchainKey}
M.~Baza, M.~Fouda, M.~Nabil, A.~S. Tag, H.~Mansour, and M.~Mahmoud,
  ``Blockchain-based distributed key management approach tailored for smart
  grid,'' in \emph{Combating Security Challenges in the Age of Big Data}.\hskip
  1em plus 0.5em minus 0.4em\relax Springer, 2019.

\bibitem{firmware2}
M.~Baza, J.~Baxter, N.~Lasla, M.~Mahmoud, M.~Abdallah, and M.~Younis,
  ``Incentivized and secure blockchain-based firmware update and dissemination
  for autonomous vehicles,'' in \emph{Connected and Autonomous Vehicles in
  Smart Cities}.\hskip 1em plus 0.5em minus 0.4em\relax CRC press, 2020.

\bibitem{tolu1}
T.~A. Odetola, H.~R. Mohammed, and S.~R. Hasan, ``A stealthy hardware trojan
  exploiting the architectural vulnerability of deep learning architectures:
  Input interception attack (iia),'' \emph{arXiv preprint arXiv:1911.00783},
  2019.

\bibitem{bacis2017pipelined}
M.~Bacis, G.~Natale, E.~Del~Sozzo, and M.~D. Santambrogio, ``A pipelined and
  scalable dataflow implementation of convolutional neural networks on fpga,''
  in \emph{2017 IEEE International Parallel and Distributed Processing
  Symposium Workshops (IPDPSW)}.\hskip 1em plus 0.5em minus 0.4em\relax IEEE,
  2017, pp. 90--97.

\bibitem{hailesellasie2019mulnet}
M.~T. Hailesellasie and S.~R. Hasan, ``{MulNet: A Flexible CNN Processor with
  Higher Resource Utilization Efficiency for Constrained Devices},'' \emph{IEEE
  Access}, 2019.

\bibitem{guo2017angel}
K.~Guo, L.~Sui, J.~Qiu, J.~Yu, J.~Wang, S.~Yao, S.~Han, Y.~Wang, and H.~Yang,
  ``{Angel-Eye: A complete design flow for mapping CNN onto embedded FPGA},''
  \emph{IEEE Transactions on Computer-Aided Design of Integrated Circuits and
  Systems}, vol.~37, no.~1, pp. 35--47, 2017.

\bibitem{park2016fpga}
J.~Park and W.~Sung, ``{FPGA based implementation of deep neural networks using
  on-chip memory only},'' in \emph{Acoustics, Speech and Signal Processing
  (ICASSP), 2016 IEEE International Conference on}.\hskip 1em plus 0.5em minus
  0.4em\relax IEEE, 2016, pp. 1011--1015.

\bibitem{rastegari2016xnor}
M.~Rastegari, V.~Ordonez, J.~Redmon, and A.~Farhadi, ``{XNOR-net: Imagenet
  classification using binary convolutional neural networks},'' in
  \emph{European Conference on Computer Vision}.\hskip 1em plus 0.5em minus
  0.4em\relax Springer, 2016, pp. 525--542.

\bibitem{zhang2017machine}
X.~Zhang, A.~Ramachandran, C.~Zhuge, D.~He, W.~Zuo, Z.~Cheng, K.~Rupnow, and
  D.~Chen, ``Machine learning on fpgas to face the iot revolution,'' in
  \emph{Proceedings of the 36th International Conference on Computer-Aided
  Design}.\hskip 1em plus 0.5em minus 0.4em\relax IEEE Press, 2017, pp.
  819--826.

\bibitem{wang2009electronic}
L.-T. Wang, Y.-W. Chang, and K.-T.~T. Cheng, \emph{Electronic design
  automation: synthesis, verification, and test}.\hskip 1em plus 0.5em minus
  0.4em\relax Morgan Kaufmann, 2009.

\bibitem{xiang2018output}
W.~Xiang, H.-D. Tran, and T.~T. Johnson, ``{Output reachable set estimation and
  verification for multilayer neural networks},'' \emph{IEEE transactions on
  neural networks and learning systems}, vol.~29, no.~11, pp. 5777--5783, 2018.

\bibitem{dwarakanath2018identifying}
A.~Dwarakanath, M.~Ahuja, S.~Sikand, R.~M. Rao, R.~Bose, N.~Dubash, and
  S.~Podder, ``Identifying implementation bugs in machine learning based image
  classifiers using metamorphic testing,'' in \emph{Proceedings of the 27th ACM
  SIGSOFT International Symposium on Software Testing and Analysis}.\hskip 1em
  plus 0.5em minus 0.4em\relax ACM, 2018, pp. 118--128.

\bibitem{mu2018collaborative}
J.~Mu, W.~Zhang, H.~Liang, and S.~Sinha, ``{A Collaborative Framework for
  FPGA-based CNN Design Modeling and Optimization},'' in \emph{2018 28th
  International Conference on Field Programmable Logic and Applications
  (FPL)}.\hskip 1em plus 0.5em minus 0.4em\relax IEEE, 2018, pp. 139--1397.

\bibitem{hao2019fpga}
C.~Hao, X.~Zhang, Y.~Li, S.~Huang, J.~Xiong, K.~Rupnow, W.-m. Hwu, and D.~Chen,
  ``{FPGA/DNN Co-Design: An Efficient Design Methodology for IoT Intelligence
  on the Edge},'' \emph{arXiv preprint arXiv:1904.04421}, 2019.

\bibitem{park2017optimizing}
H.~Park, C.~Lee, H.~Lee, Y.~Yoo, Y.~Park, I.~Kim, and K.~Yi, ``{Optimizing DCNN
  FPGA accelerator design for handwritten hangul character recognition:
  work-in-progress},'' in \emph{Proceedings of the 2017 International
  Conference on Compilers, Architectures and Synthesis for Embedded Systems
  Companion}.\hskip 1em plus 0.5em minus 0.4em\relax ACM, 2017, p.~11.

\bibitem{o2014xilinx}
D.~O'Loughlin, A.~Coffey, F.~Callaly, D.~Lyons, and F.~Morgan, ``Xilinx vivado
  high level synthesis: Case studies,'' 2014.

\bibitem{lacey2016deep}
G.~Lacey, G.~W. Taylor, and S.~Areibi, ``{Deep learning on FPGAs: Past,
  present, and future},'' \emph{arXiv preprint arXiv:1602.04283}, 2016.

\bibitem{jia2014caffe}
Y.~Jia, E.~Shelhamer, J.~Donahue, S.~Karayev, J.~Long, R.~Girshick,
  S.~Guadarrama, and T.~Darrell, ``Caffe: Convolutional architecture for fast
  feature embedding,'' in \emph{Proceedings of the 22nd ACM international
  conference on Multimedia}.\hskip 1em plus 0.5em minus 0.4em\relax ACM, 2014,
  pp. 675--678.

\bibitem{guo2016dynamic}
Y.~Guo, A.~Yao, and Y.~Chen, ``Dynamic network surgery for efficient dnns,'' in
  \emph{Advances In Neural Information Processing Systems}, 2016, pp.
  1379--1387.

\bibitem{bbb2017}
B.~Jan�en, T.~Wingender, and M.~H�bner, ``{Hardware Accelerator Framework
  Approach for Dynamic Partial Reconfigurable Overlays on Xilinx PYNQ},'' in
  \emph{INFORMATIK 2017}, M.~Eibl and M.~Gaedke, Eds.\hskip 1em plus 0.5em
  minus 0.4em\relax Gesellschaft f�r Informatik, Bonn, 2017, pp. 481--492.

\bibitem{janssen2017dynamic}
B.~Jan{\ss}en, P.~Zimprich, and M.~H{\"u}bner, ``A dynamic partial
  reconfigurable overlay concept for pynq,'' in \emph{2017 27th International
  Conference on Field Programmable Logic and Applications (FPL)}.\hskip 1em
  plus 0.5em minus 0.4em\relax IEEE, 2017, pp. 1--4.

\bibitem{pynq2019}
Xilinx, ``{Python productivity for Zynq (Pynq) Documentation Release 2.2},''
  \url{https://buildmedia.readthedocs.org/media/pdf/pynq/latest/pynq.pdf},
  2019.

\bibitem{jeff2014}
J.~Johnson, ``{Using the AXI DMA in Vivado},''
  \url{http://www.fpgadeveloper.com/2014/08/using-the-axi-dma-in-vivado.html},
  2014.

\bibitem{Xil2019}
Xilinx, ``{AXI DMA Controller},''
  \url{https://www.xilinx.com/products/intellectual-property/axi_dma.html},
  2019.

\bibitem{kastner2018hardware}
F.~K{\"a}stner, B.~Jan{\ss}en, F.~Kautz, M.~H{\"u}bner, and G.~Corradi,
  ``Hardware/software codesign for convolutional neural networks exploiting
  dynamic partial reconfiguration on pynq,'' in \emph{2018 IEEE International
  Parallel and Distributed Processing Symposium Workshops (IPDPSW)}.\hskip 1em
  plus 0.5em minus 0.4em\relax IEEE, 2018, pp. 154--161.

\bibitem{han2016eie}
S.~Han, X.~Liu, H.~Mao, J.~Pu, A.~Pedram, M.~A. Horowitz, and W.~J. Dally,
  ``{EIE: efficient inference engine on compressed deep neural network},'' in
  \emph{Computer Architecture (ISCA), 2016 ACM/IEEE 43rd Annual International
  Symposium on}.\hskip 1em plus 0.5em minus 0.4em\relax IEEE, 2016, pp.
  243--254.

\bibitem{choi2018stochastic}
J.~Choi, K.~M. Irick, J.~Hardin, W.~Qiu, A.~Yuille, J.~Sampson, and
  V.~Narayanan, ``Stochastic functional verification of dnn design through
  progressive virtual dataset generation,'' in \emph{2018 IEEE International
  Symposium on Circuits and Systems (ISCAS)}.\hskip 1em plus 0.5em minus
  0.4em\relax IEEE, 2018, pp. 1--5.

\bibitem{changwoolee}
C.~woo Lee, ``{FPGA Accelerator for CNN using Vivado HLS},''
  \url{https://github.com/changwoolee/lenet5_hls}, 2018.

\bibitem{stephendigikey}
S.~Evanczuk, ``{Get Started with Machine Learning Using Readily Available
  Hardware and Software},''
  \url{https://www.digikey.com/en/articles/techzone/2018/aug/get-started-machine-learning-hardware-and-software},
  2018.

\bibitem{alveo}
Xilinx, ``{Accelerating DNNs with Xilinx Alveo Accelerator Cards},''
  \url{https://www.xilinx.com/support/documentation/white_papers/wp504-accel-dnns.pdf},
  2018.

\end{thebibliography}

\end{document}